\documentclass[runningheads]{llncs}

\usepackage{eccv}

\usepackage{eccvabbrv}
\usepackage[breakable, skins, xparse]{tcolorbox} 
\usepackage{graphicx}
\usepackage{wrapfig}
\usepackage{algorithm}
\usepackage{algpseudocode}
\usepackage{multicol}
\usepackage{multirow}
\usepackage{subcaption} 
\usepackage{booktabs}

\usepackage{tabularx}
\usepackage{pifont}
\usepackage{makecell} 
\usepackage{amsmath}
\usepackage{amssymb}
\usepackage[inline]{enumitem}

\usepackage[table]{xcolor}
\usepackage{array}

\newcolumntype{C}[1]{>{\centering\arraybackslash}m{#1}}
\newcolumntype{Y}{>{\centering\arraybackslash}X}

\usepackage[breaklinks,colorlinks,citecolor=eccvblue]{hyperref}
\usepackage{orcidlink}

\begin{document}
\title{Video-Oasis: Rethinking Evaluation of\\ Video Understanding} 
\titlerunning{Video-Oasis: Rethinking Evaluation of Video Understanding}
\author{Geuntaek Lim\inst{1}\thanks{Work done while doing an internship at NAVER Cloud.}\orcidlink{0000-0003-2204-6109} \and 
Sungjune Park\inst{1}\orcidlink{0009-0008-0310-8718} \and 
Jaeyun Lee\inst{1}\orcidlink{0009-0009-2109-0862} \and 
Inwoong Lee\inst{2}\orcidlink{0000-0003-4356-7616} \and 
Taeoh Kim\inst{2}\orcidlink{0000-0001-7252-5525} \and 
Dongyoon Wee\inst{2}\orcidlink{0000-0003-0359-146X} \and 
Minho Shim \inst{2}\textsuperscript{$\dagger$}\orcidlink{0000-0002-9637-4909} \and 
Yukyung Choi\inst{1}\textsuperscript{$\dagger$}\orcidlink{0000-0002-9970-0132}}
\authorrunning{G. Lim et al.}
\institute{
Sejong University, South Korea \and NAVER Cloud, South Korea\\
\email{{gtlim,ykchoi}@rcv.sejong.ac.kr}
}
\maketitle
\begingroup
\renewcommand\thefootnote{$\dagger$}
\footnotetext{Co-corresponding authors.}
\endgroup

\begin{abstract}
    The inherent complexity of video understanding makes it difficult to determine whether Video-LLM benchmark performance stems from visual perception, linguistic reasoning, or knowledge priors.
    While many benchmarks have emerged to assess high-level reasoning, shared criteria for evaluating video understanding remain largely overlooked.
    Instead of introducing yet another benchmark, we take a step back to re-examine the criteria for evaluating video understanding.
    In this work, we introduce Video-Oasis, a sustainable diagnostic suite for systematically auditing existing video understanding benchmarks.
    This audit reveals that 55\% of existing benchmark samples are solvable without visual input or temporal context.
    After filtering these shortcuts, the remaining video-native challenges expose a substantial capability gap: state-of-the-art models perform only marginally above random guessing.
    Building on these findings, we use the distilled challenges as a testbed to investigate which algorithmic design choices contribute to robust video understanding.
    We hope our work provides a practical foundation for constructing rigorous video benchmarks and evaluating future Video-LLMs.
    Code is available at \href{https://github.com/sejong-rcv/Video-Oasis} {\color{magenta}{https://github.com/sejong-rcv/Video-Oasis}}.

   \vspace{-3mm}

\end{abstract}

\section{Introduction}
\label{sec:intro}
{
    \begin{figure*}[ht!]
      \centering
      \includegraphics[width=1.0\linewidth]{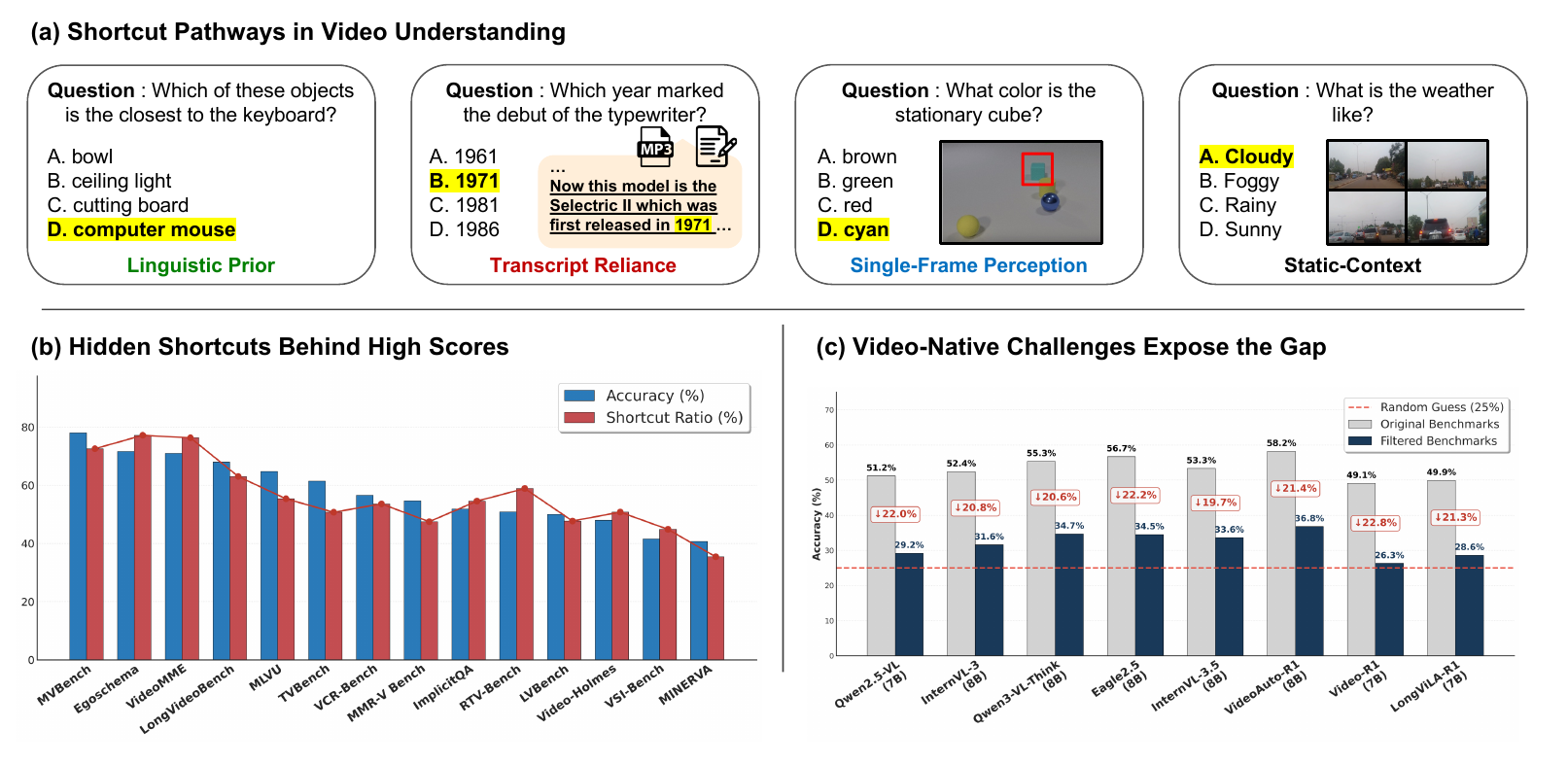}
      \caption{(a) Existing video-QA benchmarks include shortcut-solvable instances that can be answered without spatio-temporal video understanding. (b) Benchmarks with higher shortcut ratios tend to report higher accuracy. (c) State-of-the-art Video-LLMs consistently exhibit a substantial drop when facing video-native challenges, revealing the inherent difficulty of robust spatio-temporal understanding.} \vspace{-5mm}
      \label{fig:fig1}
    \end{figure*}

    The rise of multi-modal language models has steered video understanding from specific tasks toward the integration of perception and reasoning.
    While earlier benchmarks focused on narrow domains such as action recognition or temporal localization, video large language models (Video-LLMs)~\cite{videollava, qwem3_vl, longvu, eagle25, internvideo2} are now required to handle both fine-grained dynamics~\cite{mvbench, egoschema, tvbench} and long-form reasoning~\cite{longvideobench,videomme,mlvu}.
    This expansion, however, makes it difficult to determine whether reported benchmark performance stems from visual perception, linguistic reasoning, or knowledge priors.
    As benchmarks continue to proliferate across diverse tasks, a unified set of video-centric criteria becomes increasingly necessary for both benchmark creators and users.
    
    In this work, rather than introducing yet another benchmark, we take a step back to re-examine the essential criteria required for evaluating video understanding.
    As illustrated in \Cref{fig:fig1}(a), existing benchmarks often include samples that do not require visual evidence or temporal context, but can instead be solved through linguistic priors, audio cues, or static visual evidence.
    To address this issue, we introduce \textbf{Video-Oasis}, a diagnostic suite for auditing whether existing video understanding benchmarks truly require visual and temporal dependencies.
    Video-Oasis consists of three key components:
    \begin{enumerate*}[label=(\roman*)]
    \item visual-dependency tests that remove or replace visual evidence to identify samples solvable without grounded perception;
    \item temporal-dependency tests that perturb or remove temporal order to identify samples solvable without temporal reasoning; and
    \item ambiguity verification that uses human-in-the-loop inspection to identify annotation issues arising from the complexity of video data.
    \end{enumerate*}
    Together, these components provide a systematic way to filter shortcut-solvable samples and distill video-native challenges such as temporal continuity, causal interaction, and multi-event narratives.

    With this diagnostic suite, we conduct a large-scale audit of 14 diverse benchmarks~\cite{egoschema,tvbench,mvbench,longvideobench,videomme,mlvu,mmrv,lvbench,videoholmes,implicitqa,minerva,vsibench,vcrbench,rtvbench}, covering tasks from perception to reasoning and video durations from seconds to hours.
    Specifically, we decouple visual and temporal cues through a series of diagnostic tests and define the \textit{\textbf{shortcut ratio}} as the proportion of samples within a benchmark that can be solved without visual or temporal dependency.
    As shown in \Cref{fig:fig1}(b), reported benchmark accuracy is strongly correlated with shortcut prevalence, suggesting that high scores can be partially inflated by samples that do not require robust video understanding.

    \begin{table}[t]
    \centering
    \caption{Comparison with prior studies auditing benchmarks. \textbf{Evaluation Scope} denotes the number of benchmarks analyzed. \textbf{Diagnostic Coverage} indicates whether visual, temporal, and ambiguity-related diagnostic axes are considered, where $\bullet$ denotes multiple tests, $\circ$ denotes a single test, and $-$ denotes not considered. \textbf{Cross-Model Consensus} indicates whether conclusions are derived using an ensemble of models, and \textbf{Manual Verification} denotes whether human verification is performed.}
    \label{tab:main1}
    
    \setlength{\tabcolsep}{8pt} 
    \renewcommand{\arraystretch}{1.3}
    
    \resizebox{\textwidth}{!}{%
    \begin{tabular}{l c ccc cc}
    \toprule
    \multirow{2}{*}[-0.5em]{\shortstack{\textbf{Previous}\\\textbf{Work}}} & \multirow{2}{*}[-0.5em]{\shortstack{\textbf{Evaluation}\\\textbf{Scope}}} & \multicolumn{3}{c}{\textbf{Diagnostic Coverage}} & \multirow{2}{*}[-0.5em]{\shortstack{\textbf{Cross-Model}\\\textbf{Consensus}}} & \multirow{2}{*}[-0.5em]{\shortstack{\textbf{Manual}\\\textbf{Verification}}} \\ 
    \cmidrule(lr){3-5}
     & & \textbf{Vis.} & \textbf{Tem.} & \textbf{Amb.} & & \\ \midrule
    EgoTempo~\cite{egotempo}      & 2  & $-$ & $\circ$ & $-$ & $-$ & $-$ \\
    Cambrian-S~\cite{cambrian}    & 9  & $\bullet$ & $\circ$ & $-$ & $-$ & $-$ \\
    Apollo~\cite{apollo}        & 6  & $\circ$ & $\circ$ & $-$ & $-$ & $\checkmark$ \\ \midrule
    \rowcolor[gray]{0.95} 
    \textbf{Video-Oasis (Ours)} & \textbf{14} & \textbf{$\bullet$} & \textbf{$\bullet$} & \textbf{$\bullet$} & \textbf{$\checkmark$} & \textbf{$\checkmark$} \\ \bottomrule
    \end{tabular}%
    } \vspace{-4mm}
    \end{table}

    The Video-Oasis audit reveals two key findings:
    \begin{enumerate*}[label=(\roman*)]
    \item \textbf{High shortcut prevalence:} 55\% of existing benchmark samples are solvable without visual input or temporal context;
    \item \textbf{Limited performance on video-native challenges:} after filtering these shortcuts, state-of-the-art models~\cite{qwem3_vl,videoautor1,longvilar1,internvl35} perform only marginally above random chance, as shown in \Cref{fig:fig1}(c).
    \end{enumerate*}
    These findings indicate that current benchmarks can overestimate models' video understanding capabilities, while the remaining video-native challenges expose substantial limitations in current models.

    The broad audit scope and multi-axis diagnostic design of Video-Oasis distinguish it from prior benchmark-auditing studies.
    As shown in \Cref{tab:main1}, prior studies have identified important issues, including temporal shortcuts~\cite{egotempo}, perception-oriented task design~\cite{cambrian}, and benchmark redundancy~\cite{apollo}.
    Video-Oasis extends these efforts by jointly examining visual dependency, temporal dependency, and ambiguity across 14 benchmarks.
    It further incorporates cross-model consensus and human verification to improve the reliability of the diagnostic process.

    The remainder of this paper is organized as follows. We first introduce the Video-Oasis diagnostic suite and audit existing benchmarks in \Cref{sec:criteria}. We then analyze the distilled video-native challenges and evaluate a broad range of video understanding models in \Cref{sec:benchmark}. Finally, we use these challenges as a testbed to investigate algorithmic design choices for robust video understanding in \Cref{sec:method}.
    In this work, we make the following major contributions:
    \begin{itemize}
        \item \textbf{Revisiting Video Understanding.} We revisit the fundamental criteria that video understanding benchmarks should satisfy and identify a critical gap in current benchmark design.
        \item \textbf{Holistic Diagnostic Framework.} We propose Video-Oasis, a rigorous and sustainable diagnostic framework that jointly examines visual dependency, temporal dependency, and ambiguity to distill video-native challenges.
        \item \textbf{Practical Guidelines.} Through comprehensive analyses enabled by Video-Oasis, we derive practical guidelines for benchmark construction and the algorithmic design of future video understanding models.
    \end{itemize}
}

\section{Related Work}
\label{sec:related}
{
    \subsection{Large Language Models for Video Understanding}
        The evolution of Video-LLMs~\cite{videollava, qwem3_vl, longvu, eagle25, internvideo2} has centered on aligning high-dimensional visual features with the semantic space of LLMs. 
        Early research employed temporal pooling~\cite{videollava} or attention mechanisms~\cite{internvideo2} to compress video frames into token sequences within the model's context window.
        More recently, the field has shifted toward scaling context windows~\cite{longvu,eagle25, videoxl} to process longer video sequences without aggressive temporal compression.
        Despite these advancements, Video-LLMs often struggle with complex temporal narratives due to their fixed, feed-forward processing paradigm.
        To address this, dynamic agentic frameworks~\cite{videotree, worldmm, lvagent, dvd, adavideorag, vgent, HAVEN, VideoChatM1} have emerged as a complementary paradigm, utilizing structured task decomposition and iterative reasoning loops. 
        While the synergy between Video-LLMs and agentic frameworks has led to rapid gains on existing benchmarks~\cite{egoschema,longvideobench,videomme,mlvu}, it remains unclear whether these gains stem from visual perception, linguistic reasoning, or knowledge priors.
    
    \subsection{Challenges in Video Benchmark Construction}
        As Video-LLMs~\cite{eagle25,videoautor1} and agentic methods~\cite{VideoChatM1,HAVEN,videotool} rapidly improve performance on existing benchmarks~\cite{egoschema,videomme,longvideobench,mlvu}, evaluation benchmarks must be constructed with greater rigor to properly attribute these gains.
        However, the inherent complexity of video understanding, combined with the large volume of data, often makes the construction of evaluation datasets challenging.
        Consequently, dataset construction pipelines frequently rely on automatic strategies, such as generating questions from selected keyframes~\cite{Videoespresso, videochat} or using LLMs to produce questions based on video transcripts~\cite{vsibench, longvideobench, mvbench}.
        In this process, it becomes unclear whether the resulting benchmarks truly evaluate video-specific properties that distinguish video from other modalities, such as temporal continuity, causal interaction, and multi-event narratives.
        This gap calls for a rigorous re-examination of whether current pipelines preserve essential video-specific dependencies and whether reported gains reflect spatio-temporal reasoning.
            
    \subsection{Toward Robust Video Understanding Benchmarks}
        As benchmark proliferation enables comprehensive evaluation, it also introduces unintended issues. 
        Apollo~\cite{apollo} highlights the redundancy across existing benchmarks, while Cambrian-S~\cite{cambrian} observes that many tasks remain overly concentrated on perception-oriented evaluation.
        In line with these works~\cite{apollo,egotempo,cambrian}, we take a step back to re-examine the current landscape of video understanding.
        Building upon these efforts, we introduce Video-Oasis, a diagnostic suite that distills existing datasets to isolate core spatio-temporal challenges while suppressing unintended shortcuts through visual–temporal decoupling tests.
        By integrating cross-model consensus and human-in-the-loop verification, Video-Oasis extends prior analyses by providing a sustainable framework for auditing modern video understanding benchmarks.
}

\section{Video-Oasis: Diagnostic Suite for Video Understanding}
\label{sec:criteria} 
{   
    \begin{figure*}[t!]
      \centering
      \includegraphics[width=0.98\linewidth]{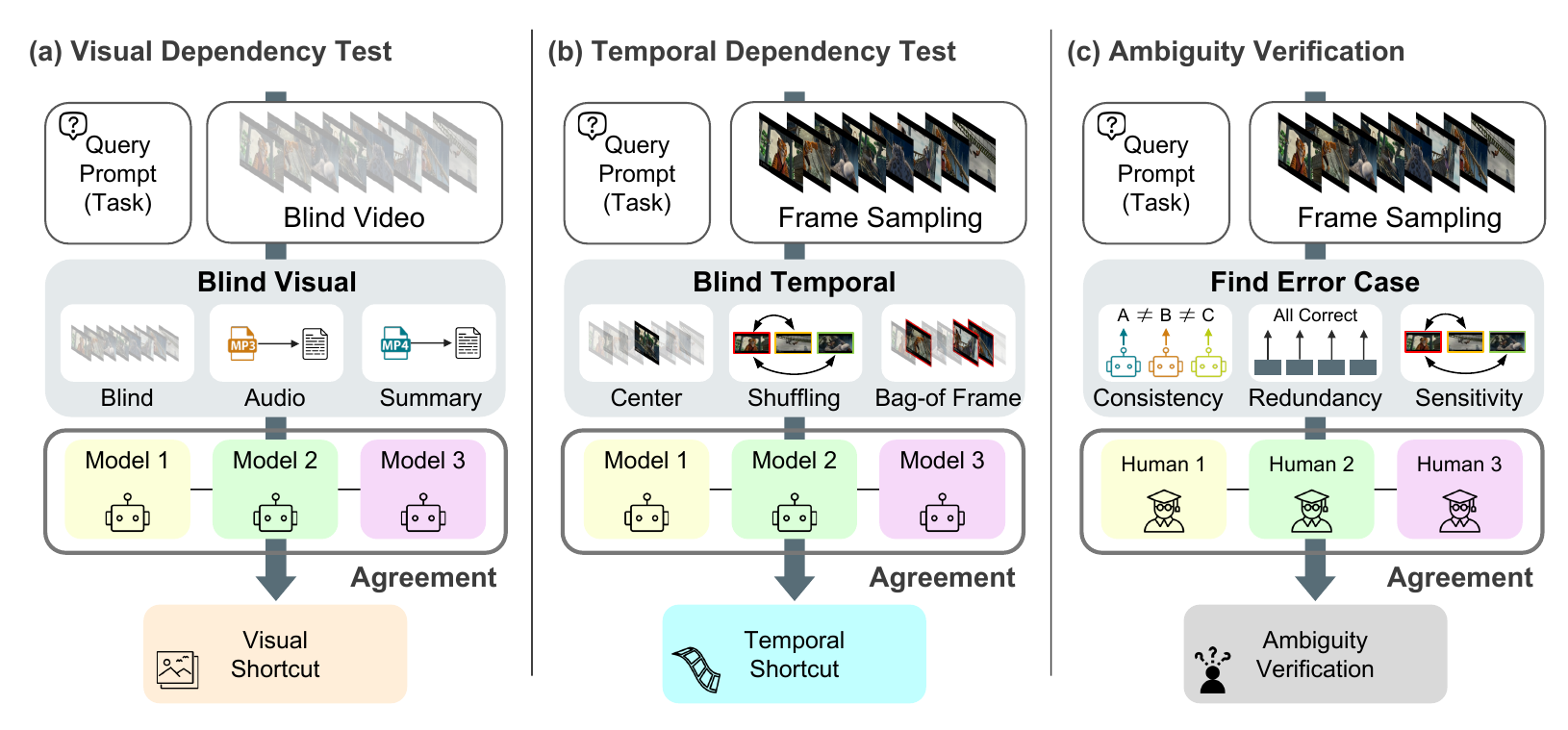}
      \caption{Overview of the Video-Oasis diagnostic suite, which assesses (a) whether visual information is required, (b) whether temporal context is necessary, and (c) whether the task contains ambiguity in video data.}
      \vspace{-4mm}
      \label{fig:fig2}
    \end{figure*}

    Establishing reliable protocols for measuring spatio-temporal reasoning remains a critical yet underexplored challenge in video understanding.
    To address this, we introduce Video-Oasis, a diagnostic suite for verifying whether video benchmarks satisfy shared criteria for video understanding.
    First, we introduce our test design, which systematically examines the essential dependencies required for video understanding (\Cref{sec:3.1}).
    Next, we audit existing benchmarks using our diagnostic protocols (\Cref{sec:3.2}).
    Finally, we examine the diagnostic coverage of Video-Oasis and the validity of shortcut identification (\Cref{sec:3.3}).
    Experimental settings are detailed in Sec. C of the supplementary material. \vspace{-4mm}

    \subsection{Design of the Diagnostic Suite}
    \label{sec:3.1}
        \textbf{Criteria 1: Is Visual Evidence Required?}
        To test visual dependency, we replace the original video with inputs that remove or abstract away raw visual evidence.
        As illustrated in \Cref{fig:fig2}(a), we use three diagnostic tests: (i) \textbf{Blind}, which provides only the question and answer options to identify cases solvable through linguistic bias or world knowledge without any visual input; (ii) \textbf{Audio}, where the video’s audio track is transcribed into text and given to the model in place of the video; and (iii) \textbf{Summary}, where the raw video is replaced by a concatenated sequence of captions~\cite{care} extracted at fixed intervals.
        
        The Audio and Summary tests are intentionally simple and are not designed to optimize text-only video reasoning. 
        Although recent methods have explored treating video reasoning as long-document comprehension through iterative retrieval or memory updating~\cite{drvideo,videotree}, we use these textual inputs for a different purpose.
        They serve as diagnostic probes: if a sample can be answered from an audio transcript or caption sequence, it may not require grounded visual perception from the raw video.
        
        \begin{figure*}[ht!]
          \centering
          \includegraphics[width=0.9\linewidth]{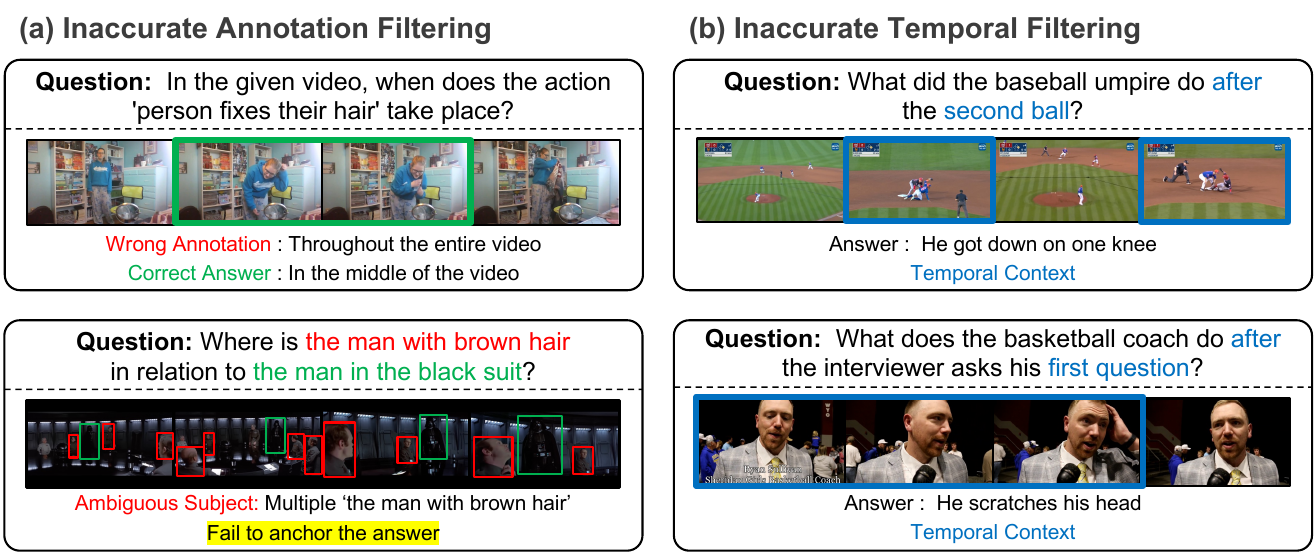}
          \caption{(a) Inaccurate annotations identified by the redundancy and consistency tests. (b) Questions incorrectly filtered by the frame shuffling test but manually restored.} \vspace{-6mm}
          \label{fig:fig3}
        \end{figure*}
        
        \par\smallskip
        \noindent\textbf{Criteria 2: Is Temporal Context Required?}
        Temporal dependency is central to video understanding because many questions require ordering events, tracking state changes, or reasoning about causal interactions.
        As illustrated in \Cref{fig:fig2}(b), we verify this dependency through three diagnostic tests: (i) \textbf{Center-Frame}, which provides only the middle frame to detect if the task functions merely as spatial recognition without temporal depth; (ii) \textbf{Frame Shuffling}, where we randomly permute the frame order to disrupt temporal causality and measure the model's sensitivity to chronological sequences; and (iii) \textbf{Bag-of-Frames (BoF)}, which employs a frozen CLIP-based encoder~\cite{clip,longclip,evaclip} that does not model temporal order to perform top-$k$ frame matching against the query.
        If this non-temporal, similarity-based approach succeeds, the task does not necessitate temporal reasoning but rather simple visual pattern recognition. 
        
        \par\smallskip
        \noindent\textbf{Criteria 3: Is the Annotation Reliable?}
        Since video data are long and information-dense, video-QA annotations can be ambiguous, involving imprecise temporal grounding, incomplete evidence, or non-unique answers.
        While many existing dataset construction pipelines often overlook this uncertainty, we apply three checks that flag samples for manual inspection, as illustrated in \Cref{fig:fig2}(c): (i) \textbf{Consistency}, identifying cases where models~\cite{eagle25,qwem3_vl,internvl35,videoautor1,videor1} fail to reach a consensus, which suggests inherent ambiguity or non-unique answers; (ii) \textbf{Redundancy}, investigating cases solvable via any arbitrary video segment, revealing flawed question designs or global biases that fail to anchor the answer to a specific temporal segment; and (iii) \textbf{Sensitivity}, where we manually verify cases in which models succeed despite frame shuffling to account for potential uncertainty in the sequential understanding of Video-LLMs.

        \Cref{fig:fig3} provides representative examples of the manual verification process.
        \Cref{fig:fig3}(a) shows annotation issues identified by the consistency and redundancy checks, including incorrect temporal labels and ambiguous subjects that prevent the answer from being reliably anchored to the video.
        \Cref{fig:fig3}(b) shows cases initially filtered by the frame-shuffling test, but manually restored because the questions still require temporal ordering, such as reasoning about what happens after a specific event.

    \subsection{Benchmarking the Benchmarks}
    \label{sec:3.2}
        \noindent\textbf{Diagnostic Test Results.} 
        We conduct comprehensive diagnostic tests across existing benchmarks, leveraging a diverse set of Video-LLMs~\cite{eagle25,qwem3_vl,qwen25vl,videoautor1} and VLMs~\cite{clip,longclip,evaclip}.
        The results in \Cref{tab:main2} are aggregated over all 14 benchmarks.
        Notably, models achieve accuracies ranging from 30\% to 50\% even when raw visual evidence or temporal order is removed or disrupted, compared to the random-chance baseline of 25.6\%.
        These results suggest that many benchmark samples do not strictly enforce the intended visual or temporal dependencies.
        Additional robustness analysis and benchmark-wise results are provided in Secs. A and B of the supplementary material, respectively.
                                    
        \begin{table}[h]
            \centering
            \vspace{-4mm}
            \footnotesize 
            \setlength{\tabcolsep}{5pt} 
            \begin{subtable}[t]{0.31\textwidth}
                \centering
                \begin{tabular}{lc}
                    \toprule
                    \rowcolor[gray]{0.9} \textbf{Video-LLM} & \textbf{Acc.} \\
                    \midrule
                    Eagle2.5 & 35.6 \\
                    Qwen2.5-VL & 33.5 \\
                    Qwen3-VL & 36.2 \\
                    \bottomrule
                \end{tabular}
                \caption{Blind Test}
            \end{subtable}
            \hfill
            \begin{subtable}[t]{0.31\textwidth}
                \centering
                \begin{tabular}{lc}
                    \toprule
                    \rowcolor[gray]{0.9} \textbf{Video-LLM} & \textbf{Acc.} \\
                    \midrule
                    Eagle2.5 & 47.6 \\
                    Qwen2.5-VL & 46.8 \\
                    Qwen3-VL & 45.9 \\
                    \bottomrule
                \end{tabular}
                \caption{Audio}
            \end{subtable}
            \hfill
            \begin{subtable}[t]{0.31\textwidth}
                \centering
                \begin{tabular}{lc}
                    \toprule
                    \rowcolor[gray]{0.9} \textbf{Video-LLM} & \textbf{Acc.} \\
                    \midrule
                    Eagle2.5 & 44.0 \\
                    Qwen2.5-VL & 42.6 \\
                    Qwen3-VL & 45.0 \\
                    \bottomrule
                \end{tabular}
                \caption{Summary}
            \end{subtable}
        
            \vspace{0.2em} 
        
            \begin{subtable}[t]{0.31\textwidth}
                \centering
                \begin{tabular}{lc}
                    \toprule
                    \rowcolor[gray]{0.9} \textbf{Video-LLM} & \textbf{Acc.} \\
                    \midrule
                    Eagle2.5 & 42.2 \\
                    Qwen3-VL & 40.2 \\
                    VideoAuto-R1 & 43.0 \\
                    \bottomrule
                \end{tabular}
                \caption{Center-Frame}
            \end{subtable}
            \hfill
            \begin{subtable}[t]{0.31\textwidth}
                \centering
                \begin{tabular}{lc}
                    \toprule
                    \rowcolor[gray]{0.9} \textbf{Video-LLM} & \textbf{Acc.} \\
                    \midrule
                    Eagle2.5 & 52.2 \\
                    Qwen3-VL & 50.7 \\
                    VideoAuto-R1 & 52.4 \\
                    \bottomrule
                \end{tabular}
                \caption{Frame Shuffling}
            \end{subtable}
            \hfill
            \begin{subtable}[t]{0.31\textwidth}
                \centering
                \begin{tabular}{lc}
                    \toprule
                    \rowcolor[gray]{0.9} \textbf{VLM} & \textbf{Acc.} \\
                    \midrule
                    $\text{CLIP}$ & 31.4 \\
                    $\text{Long-CLIP}$ & 32.8 \\
                    $\text{EVA-CLIP}$ & 32.7 \\
                    \bottomrule
                \end{tabular}
                \caption{Bag-of-Frames}
            \end{subtable}
            \caption{Aggregate diagnostic-test results over 14 benchmarks. The metric is accuracy.} \vspace{-10mm}
            \label{tab:main2}
        \end{table}
    
        \noindent\textbf{Auditing Existing Benchmarks.}
        To conduct a granular analysis based on task characteristics, we manually group the 14 benchmarks into spatial, temporal, reasoning, and general categories.
        We define the consensus threshold ($c$) as the number of diagnostic models listed in \Cref{tab:main2} that must answer a sample correctly for it to be counted as a shortcut.
        Under this framework, we aggregate shortcut instances across all tests and compare their ratios under different consensus thresholds.
        \Cref{tab:main3} reveals two key findings: (i) shortcut-solvable samples appear consistently across all task groups, and (ii) under a relaxed consensus threshold ($c\geq1$), an average of 92.7\% of samples exhibit shortcut-solvable behavior under at least one visual or temporal diagnostic test.
        These results indicate that many existing benchmarks do not sufficiently enforce the spatio-temporal dependencies required for video understanding.
        
        \begin{table}[h]
        \vspace{-6mm}
        \caption{Ratio of shortcuts (\%) across comprehensive benchmark groups.} 
        \centering
        \label{tab:main3}
        \small
        \setlength{\tabcolsep}{6pt}
            \begin{tabular}{ccccc}
                \toprule
                \makecell{\textbf{Consensus} \\ \textbf{Threshold}} & 
                \makecell{\textbf{Spatial} \\ \cite{egoschema,vsibench,implicitqa}} & 
                \makecell{\textbf{Temporal} \\ \cite{rtvbench,tvbench,vcrbench}} & 
                \makecell{\textbf{Reasoning} \\ \cite{minerva,videoholmes,mmrv}} & 
                \makecell{\textbf{General} \\ \cite{longvideobench,mlvu,videomme,mvbench,lvbench}} \\
                \midrule
                $ c\geq 1$ & 95.6 & 95.7 & 85.8 & 94.0 \\
                $ c\geq 2$ & 86.1 & 85.2 & 69.2 & 83.9 \\
                $c=3$ & 58.8 & 54.4 & 44.6 & 63.0 \\
                \bottomrule
        \end{tabular} 
        \end{table}

    \subsection{Empirical Insights into the Diagnostic Framework}
    \label{sec:3.3}
        \noindent\textbf{Diagnostic Test Distribution.}
        Video-Oasis includes ambiguity tests to refine unreliable or uncertain samples before constructing the final filtered set.
        As summarized in \Cref{tab:main4}, the consistency and redundancy checks address unreliable annotations, while the sensitivity check corrects potential false positives from frame shuffling tests.
        We then analyze how shortcuts are distributed across the individual diagnostic tests of Video-Oasis.
        Under the strict consensus condition ($c=3$), \Cref{tab:main5} reports both total and unique shortcut counts for each diagnostic test.
        Since the tests are not perfectly orthogonal, the unique counts measure the distinct contribution of each test beyond overlaps with others.
        \Cref{tab:main5} reveals two findings: (i) \textit{Summary}, \textit{Center-Frame}, and \textit{Frame Shuffling} account for the majority of unique shortcuts (65\%), forming a practical protocol for future benchmark construction; and (ii) \textit{Blind}, \textit{Audio}, and \textit{Bag-of-Frames} still contribute a non-negligible 35\%, demonstrating that Video-Oasis provides complementary diagnostic coverage.

        \begin{table}[h]
            \vspace{-6mm}
            \centering
            \renewcommand{\arraystretch}{1.1} 
            \setlength{\tabcolsep}{8pt} 
            \begin{minipage}[t]{0.48\linewidth}
                \centering
                \caption{Statistics of manual refinement.}
                \label{tab:main4}
                \resizebox{0.95\linewidth}{!}{ 
                    \begin{tabular}{lcc}
                        \toprule
                        \multirow{2}{*}{Ambiguity Test} & \multicolumn{2}{c}{\# of Samples} \\
                        \cmidrule(lr){2-3}
                         & Total & Refined \\
                        \midrule
                        Consistency & 666 & 213 \\
                        Redundancy & 477 & 197 \\
                        Sensitivity & 1,758 & 804 \\
                        \bottomrule
                    \end{tabular}
                }
            \end{minipage}
            \hfill
            \begin{minipage}[t]{0.48\linewidth}
                \centering
                \caption{Statistics of diagnostic tests.}
                \label{tab:main5}
                \resizebox{0.95\linewidth}{!}{ 
                    \begin{tabular}{lcc}
                        \toprule
                        \multirow{2}{*}{Diagnostic Test} & \multicolumn{2}{c}{\# of Samples} \\
                        \cmidrule(lr){2-3} 
                         & Total & Unique \\
                        \midrule
                        Blind & 2,751 & 362 \\
                        Audio & 1,685 & 301 \\
                        Summary & 6,703 & 1,308 \\
                        Center-Frame & 5,882 & 847 \\
                        Frame Shuffling & 8,280 & 1,758 \\
                        Bag-of-Frames & 4,309 & 1,394 \\
                        \bottomrule
                    \end{tabular}
                }
            \end{minipage}
            \vspace{-5mm}
        \end{table}
        
        \noindent\textbf{Validity of Shortcut Identification.} 
        Because shortcut identification relies on restricted settings that remove or weaken visual or temporal evidence, success in these settings may not verify shortcut behavior.
        To assess the validity of shortcut identification, we define the correlation rate as the fraction of shortcut-identified samples that are correctly solved under standard evaluation with uniformly sampled frames.
        We use different models~\cite{internvl35,longvilar1,videotool} from those used for shortcut identification to avoid circular validation.
        As \Cref{tab:main6} shows, the identified samples exhibit a high correlation rate, averaging 76\% across tests and models.
        This indicates that the identified shortcut cases are already handled well by current models, suggesting that future evaluation should move beyond them and focus on the remaining problems that continue to challenge current models.
        \vspace{-2mm}

        \begin{table}[ht]
        \vspace{-4mm}
        \centering
        \caption{Correlation rate (\%) of shortcut-identified samples under standard evaluation.}
        \label{tab:main6}
        \small
        \setlength{\tabcolsep}{4pt}
            \resizebox{0.95\linewidth}{!}{
            \begin{tabular}{l cccccc}
                \toprule
                \textbf{Models} & \textbf{Blind} & \textbf{Audio} & \textbf{Summary} & 
                \makecell{\textbf{Center} \\ \textbf{Frame}} & 
                \makecell{\textbf{Bag-of} \\ \textbf{Frames}} & 
                \makecell{\textbf{Frame} \\ \textbf{Shuffling}} \\
                \midrule
                InternVL-3.5 (8B)~\cite{internvl35} & 77.0 & 74.0 & 79.2 & 79.4 & 67.2 & 84.1 \\
                LongViLA-R1 (7B)~\cite{longvilar1} & 78.8 & 72.8 & 78.1 & 77.7 & 65.7 & 81.3 \\
                STAR~\cite{videotool} & 74.0 & 79.8 & 76.4 & 74.5 & 68.0 & 76.5 \\
                \bottomrule
            \end{tabular}} \vspace{-4mm}
        \end{table}
}

\section{Distilling the Challenges of Video Understanding}
\label{sec:benchmark}
{
    We next examine the challenges that remain after filtering shortcut-solvable samples and evaluate how state-of-the-art models perform on these video-native challenges.
    Specifically, we first identify the types of video-native challenges distilled by Video-Oasis (\Cref{sec:4.1}), and then evaluate state-of-the-art models under this distilled setting to reveal the remaining gap in strict spatio-temporal understanding (\Cref{sec:4.2}). \vspace{-4mm}

    \subsection{Understanding the Video-Native Challenges}    
    \label{sec:4.1}
        After applying Video-Oasis, the remaining samples are more likely to require strong visual and temporal dependencies.
        We further analyze these samples to understand what types of video-native challenges they represent.
        To this end, rather than manually inspecting each sample, we design an efficient pipeline that leverages existing annotations.

        Specifically, we first aggregate the source-benchmark metadata associated with the surviving QA pairs, including their original task categories and sub-task labels.
        Using this metadata, we prompt Gemini-2.5-Pro~\cite{gemini25} to derive candidate challenge clusters by abstracting the remaining tasks under the Video-Oasis criteria.
        We then consolidate these initial clusters into five unified categories that capture the dominant capabilities required by the Video-Oasis-filtered samples.
        
        \begin{itemize}
            \item \textbf{Fine-Grained Perception:} requires grounding fine-grained recognition in a spatio-temporal context, where visual identification depends on how details evolve across space and time.
            
            \item \textbf{Spatial World Understanding:} requires synthesizing fragmented, multi-view evidence across frames to infer 3D context, including relative positions, geometry, and trajectories.
            
            \item \textbf{Temporal Dynamics \& Tracking:} requires monitoring changes over time, such as object tracking, action sequencing, and state transitions, demanding temporal ordering to prevent reliance on unordered frame matching.

            \item \textbf{Causality \& Logical Reasoning:} requires deducing latent cause-and-effect relationships, physical laws, and unobserved intentions, going beyond pixel-level observations to probe the implicit logic of the video.
            
            \item \textbf{Global Narrative:} requires integrating events across the full timeline to infer long-term semantics or overarching plots while filtering out irrelevant contexts over extended video horizons.
        \end{itemize}

        Next, we assign each surviving QA pair to a single primary challenge category.
        For this categorization step, each annotator model receives the question, answer options, and the definitions of the five categories.
        To improve annotation consistency, we employ an ensemble of five proprietary LLMs~\cite{gpt4o,gpt5,o4mini}.
        A category label is accepted when at least three models agree; otherwise, the sample is manually inspected and labeled.
        Only 122 samples fail to reach consensus among the five LLMs and require manual inspection.
        Further details on category annotation prompts and robustness to annotation-model choices are provided in Sec. D.3 of the supplementary material.

        Our goal is not to introduce novel tasks, but to refine the fundamental capabilities of video understanding. 
        While these categories may resemble traditional taxonomies, they are derived from a bottom-up, data-driven process. 
        By filtering out shortcut-solvable samples, these video-native challenges are derived from the remaining tasks rather than imposed as predefined benchmark categories.
        
        \begin{figure*}[t!]
          \centering
          \includegraphics[width=1.0\linewidth]{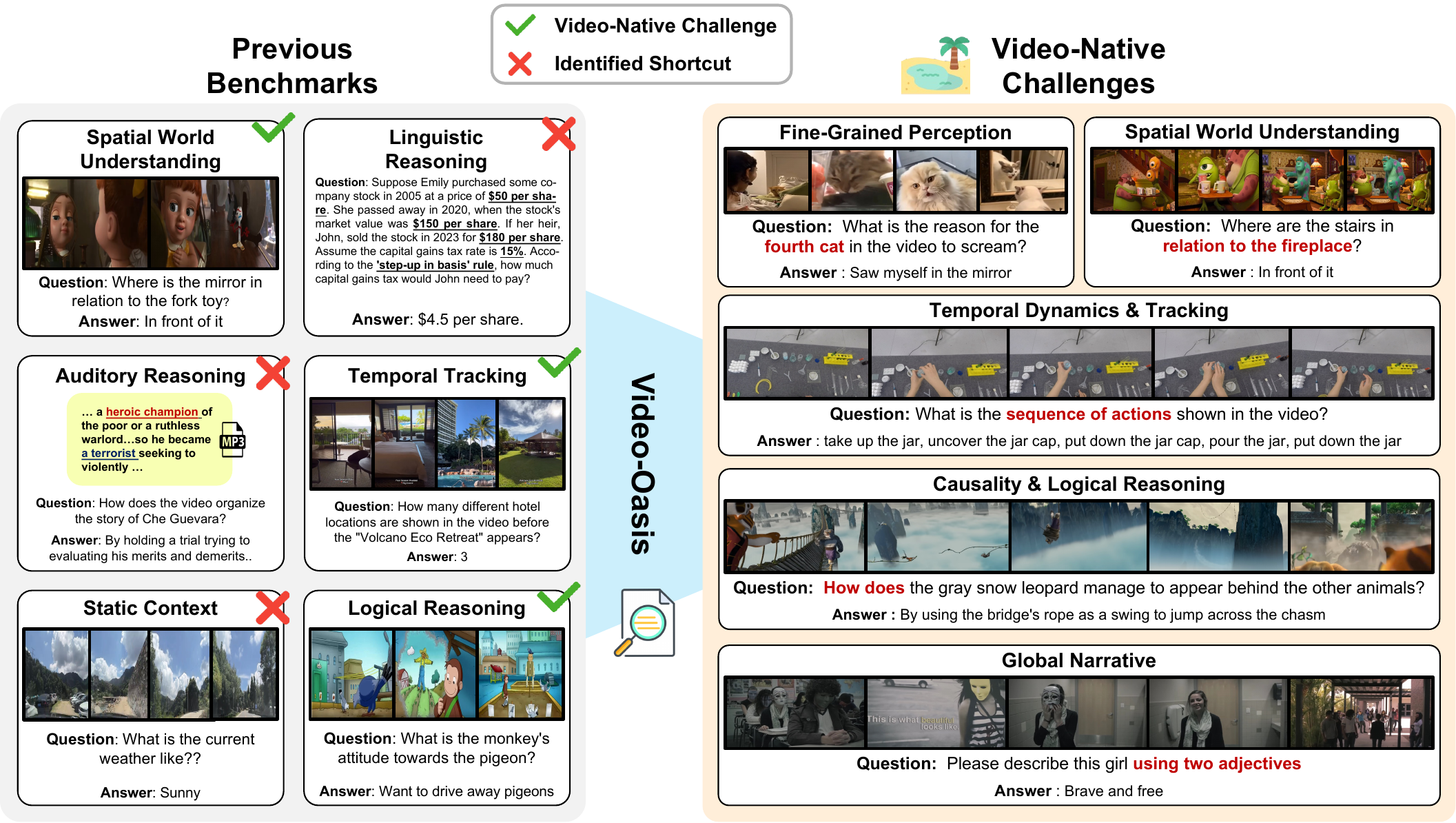}
          \caption{Video-Oasis filters shortcut-driven samples from existing benchmarks and distills video-native challenges that require fine-grained perception, spatial understanding, temporal tracking, causal reasoning, and global narrative understanding.}
          \label{fig:fig4} \vspace{-6mm}
        \end{figure*}

        \par\smallskip
        \noindent \textbf{Overview.} 
        From the 14 curated benchmarks covering diverse aspects of video understanding, 11,033 QA pairs remain from the original 24,416, associated with 4,938 unique videos. 
        By reducing the evaluation volume by 55\%, Video-Oasis enables more efficient evaluation while preserving essential spatio-temporal challenges. 
        The diagnostic tests used to construct this set are reproducible and extensible, enabling seamless incorporation of new benchmarks and emerging challenges. 
        Additional statistics on the distilled video-native challenges are provided in Sec. D.1 of the supplementary material.

        \par\smallskip
        \noindent \textbf{Qualitative Examples.} 
        Video-Oasis establishes a rigorous evaluation setting that specifically targets strict spatio-temporal dependencies.
        In \Cref{fig:fig4}(left), tasks solvable through simple frame-level perception or linguistic/auditory reasoning without visual input fail to satisfy video-specific criteria.
        Moving beyond simple frame-level recognition, Video-Oasis highlights challenges such as spatial and semantic matching across disparate views, chronological reasoning, and the preservation of temporal continuity, as illustrated in \Cref{fig:fig4}(right).
        These scenarios represent inherently video-native challenges that remain difficult to solve without access to visual and temporal evidence, highlighting the core directions that future video understanding evaluation should emphasize.
        More qualitative examples and category-level explanations are provided in Sec. D.2 of the supplementary material.

        \noindent\textbf{Beyond Difficulty-Based Filtering.} 
        We further examine whether the distilled video-native challenges are merely difficult questions for current models. 
        To this end, we compare the Video-Oasis-distilled set with a difficulty-based baseline of 6,609 questions that Eagle2.5~\cite{eagle25}, Qwen3-VL~\cite{qwem3_vl}, and VideoAuto-R1~\cite{videoautor1} all answer incorrectly.
        The two sets show only 44.6\% overlap.
        This indicates that Video-Oasis does not simply collect hard questions, but identifies samples through predefined visual, temporal, and ambiguity diagnostics that better reflect video-native dependencies.
        
    \subsection{Comprehensive Evaluation}
    \label{sec:4.2}
        \noindent \textbf{Experimental Settings.} 
        We conduct an extensive evaluation across a diverse spectrum of models, encompassing open-source Video-LLMs~\cite{qwem3_vl,eagle25,videoautor1,videor1}, leading proprietary models~\cite{gpt4o,gemini25}, and agentic methods~\cite{videotool,videotree}.
        To ensure a fair comparison across all models, the input visual context is restricted to a maximum of 128 frames, uniformly sampled at a rate of 1 fps.
        For agentic frameworks~\cite{videotree,videotool}, we adhere to the default configurations in their original implementations, except that the reasoning model is replaced with GPT-5-mini~\cite{gpt5}.
        To ensure experimental reliability, we provide a detailed validation comparing official benchmark scores with our reproduced results in Sec. E of the supplementary material.

        \begin{table}[ht]
            \centering
            \caption{Benchmarking state-of-the-art models under video-native challenges.}
            \label{tab:main7}
            \small
            \setlength{\tabcolsep}{3pt} 
            \resizebox{\columnwidth}{!}{
                \begin{tabular}{l ccccc c}
                    \toprule
                    \textbf{Models} & \makecell{\textbf{Fine.} \\ \textbf{Percep.}} & \makecell{\textbf{Spatial} \\ \textbf{World}} & \makecell{\textbf{Temporal} \\ \textbf{Dynamics}} & \makecell{\textbf{Causal} \\ \textbf{Reason.}} & \makecell{\textbf{Global} \\ \textbf{Narrat.}} & 
                    \textbf{Overall} \\
                    \midrule
                    \rowcolor[gray]{0.9} \multicolumn{7}{l}{\textit{Proprietary LLMs}} \\
                    GPT-4o~\cite{gpt4o} & 25.6 & 33.2 & 26.3 & 27.3 & 26.5 & 27.5 \\
                    Gemini-2.5-Pro~\cite{gemini25} & \textbf{40.2} & \textbf{49.8} & \textbf{50.9} & \textbf{45.4} & \textbf{43.0} & \textbf{46.7} \\
                    \midrule
                    \rowcolor[gray]{0.9} \multicolumn{7}{l}{\textit{Open-Source Video-LLMs}} \\
                    Qwen2.5-VL (7B)~\cite{qwen25vl} & 23.3 & 28.7 & 32.3 & 28.6 & 21.2 & 29.2 \\
                    $\text{Qwen3-VL}_\text{inst.} \text{(8B)}$~\cite{qwem3_vl} & 27.0 & 42.4 & 36.5 & 28.0 & 21.5 & 33.8 \\
                    $\text{Qwen3-VL}_\text{think.} \text{(8B)}$~\cite{qwem3_vl} & 29.0 & 41.6 & 37.7 & 27.7 & 23.2 & 34.6 \\
                    Eagle2.5 (8B)~\cite{eagle25} & 26.9 & 31.0 & 39.7 & 33.2 & 22.7 & 34.5 \\
                    InternVL-3 (8B)~\cite{internvl3} & 27.0 & 31.3 & 34.1 & 30.6 & 24.5 & 31.6 \\
                    InternVL-3.5 (8B)~\cite{internvl35} & 29.5 & 41.9 & 35.1 & 29.8 & 23.3 & 33.6 \\
                    Video-R1 (7B)~\cite{videor1} & 24.0 & 24.0 & 29.1 & 27.3 & 18.4 & 26.3 \\
                    LongViLA-R1 (7B)~\cite{longvilar1} & 28.4 & 25.4 & 31.5 & 27.9 & 20.6 & 28.6 \\
                    VideoAuto-R1 (8B)~\cite{videoautor1} & 27.5 & 44.3 & 39.5 & 31.1 & 28.9 & 36.8 \\
                    \midrule
                    \rowcolor[gray]{0.9} \multicolumn{7}{l}{\textit{Agentic Methods}} \\
                    VideoTree ($\text{GPT-5}_\text{{mini}}$)~\cite{videotree} & 28.6 & 34.0 & 32.3 & 24.6 & 20.7 & 30.1 \\
                    STAR ($\text{GPT-5}_\text{{mini}}$)~\cite{videotool} & \underline{31.6} & \underline{44.4} & \underline{42.2} & \underline{34.0} & \underline{32.9} & \underline{39.5} \\
                    \bottomrule
                \end{tabular}
            } \vspace{-4mm}
        \end{table}
                
        \noindent \textbf{Experimental Results.} 
        Comprehensive quantitative results for all methods under the distilled evaluation setting are shown in \Cref{tab:main7}.
        We report category-wise accuracy for each model and use aggregate accuracy over all samples as the overall score.
        From these results, we draw several conclusions:
        \begin{enumerate}
            \item \textbf{Performance nearing random-chance levels:} A key observation is that current Video-LLMs~\cite{qwen25vl,videor1,longvilar1} exhibit performance close to the chance level (25.6\%), with the exception of a few top-tier models~\cite{videotool,gemini25}, confirming that current models struggle with rigorous spatio-temporal reasoning.
        
            \item \textbf{The current frontier of video reasoning:} Gemini-2.5-Pro~\cite{gemini25} achieves the highest performance across all skills; however, its moderate performance also indicates that these tasks pose a non-trivial challenge even for the most advanced proprietary models.
        
            \item \textbf{Bottleneck in holistic understanding:} The consistently low performance in the \textit{Global Narrative} dimension reveals that long-term multi-scene understanding remains a primary bottleneck for current architectures.
        
            \item \textbf{Impact of agentic designs:} Agentic design choices, such as VideoTree~\cite{videotree} and STAR~\cite{videotool}, significantly influence overall efficacy even when using the same reasoning model~\cite{gpt5}, underscoring the importance of reasoning-step orchestration.
                        
        \end{enumerate}

}
\section{Exploring Algorithmic Designs}
\label{sec:method}
{    
    The distilled video-native challenges expose a substantial gap in current Video-LLMs.
    We next use this setting to examine which algorithmic designs can help close this gap.
    Because these challenges emphasize visual and temporal dependencies, they provide a focused testbed for analyzing model improvements.
        
    \subsection{The Role of Temporal Grounding}
    \label{sec:5.1}
        Temporal grounding associates a query with the video segments needed to answer it, making it especially critical for video-native challenges with strict spatio-temporal dependencies.
        To investigate its practical impact, we conduct an ablation study using AKS~\cite{aks} to retrieve the 16 most relevant frames as a representative temporal grounding method.        
        As shown in \Cref{tab:main8}, integrating this grounding pipeline yields consistent improvements across models. 
        However, the magnitude of these gains remains modest, with improvements of 1.4\% for Eagle2.5~\cite{eagle25} and 2.3\% for Qwen3-VL-Instruct~\cite{qwem3_vl}, despite their relatively low baseline performance.
        This raises a natural question: are the limited gains caused by weak reasoning, or by imperfect temporal grounding?
        
        \begin{table}[ht]
            \vspace{-4mm}
            \centering
            \caption{Ablation study of temporal grounding in Video-LLMs. The metric is accuracy.}
            \label{tab:main8}
            \setlength{\tabcolsep}{4pt} 
            \resizebox{0.95\textwidth}{!}{%
            \begin{tabular}{lccccccc}
            \toprule
            \textbf{Method} & 
            \textbf{\small \begin{tabular}[c]{@{}c@{}}Temporal\\ Grounding\end{tabular}} & 
            \textbf{\small \begin{tabular}[c]{@{}c@{}}Fine.\\ Perception\end{tabular}} & 
            \textbf{\small \begin{tabular}[c]{@{}c@{}}Spatial\\ World\end{tabular}} & 
            \textbf{\small \begin{tabular}[c]{@{}c@{}}Temporal\\ Dynamics\end{tabular}} & 
            \textbf{\small \begin{tabular}[c]{@{}c@{}}Causal\\ Logical\end{tabular}} & 
            \textbf{\small \begin{tabular}[c]{@{}c@{}}Global\\ Narrative\end{tabular}} & 
            \textbf{Overall} \\ \midrule
            \multirow{2}{*}{Eagle2.5~\cite{eagle25}} & - & 25.0 & 29.4 & 34.9 & 30.5 & 25.7 & \underline{31.5} \\
                                      & \textbf{$\checkmark$} & 28.4 & 31.9 & 35.7 & 30.4 & 27.5 & \textbf{32.9} \\ \midrule
            \multirow{2}{*}{$\text{Qwen3-VL}$-$\text{Instruct}$~\cite{qwem3_vl}} & - & 22.9 & 37.1 & 28.8 & 24.8 & 16.9 & \underline{27.8} \\
                                               & \textbf{$\checkmark$} & 25.2 & 37.9 & 32.3 & 23.3 & 19.9 & \textbf{30.1} \\ \bottomrule
            \end{tabular}%
            }
            \vspace{-4mm}
        \end{table}

        To isolate the impact of grounding from uncertainty, we conducted an oracle experiment on 2,945 QA pairs from ImplicitQA~\cite{implicitqa} and KFS-Bench~\cite{kfs}.
        The set contains 1,060 Video-Oasis-distilled samples and 1,885 shortcut samples, each paired with ground-truth temporal regions.
        \Cref{tab:main9} reveals a clear contrast: while oracle grounding leads to a substantial performance gain on Video-Oasis-distilled samples ($35.0\% \rightarrow 50.8\%$), the gain in shortcuts is notably smaller ($78.0\% \rightarrow 80.8\%$), confirming that samples identified as shortcuts often allow models to bypass precise temporal grounding. 
        This finding yields a crucial insight: \textbf{precise grounding becomes increasingly important in environments where strong spatio-temporal dependencies are required.} 
    
        \begin{table}[h]
            \vspace{-4mm}
            \centering
            \caption{Upper bound performance (\%) with oracle temporal grounding.}
            \label{tab:main9}
            \setlength{\tabcolsep}{4pt}
            \resizebox{1.0\textwidth}{!}{%
            \begin{tabular}{lccccccc}
            \toprule
            \textbf{Method} & \textbf{\small\begin{tabular}[c]{@{}c@{}}Fine.\\Perception\end{tabular}} & \textbf{\small\begin{tabular}[c]{@{}c@{}}Spatial\\World\end{tabular}} & \textbf{\small\begin{tabular}[c]{@{}c@{}}Temporal\\Dynamics\end{tabular}} & \textbf{\small\begin{tabular}[c]{@{}c@{}}Causal\\Logical\end{tabular}} & \textbf{\small\begin{tabular}[c]{@{}c@{}}Global\\Narrative\end{tabular}} & \textbf{\small\begin{tabular}[c]{@{}c@{}}Video-Oasis\\(overall)\end{tabular}} & \textbf{\small\begin{tabular}[c]{@{}c@{}}Shortcut\\(overall)\end{tabular}} \\ \midrule
            Eagle2.5 & 37.2 & 22.9 & 40.5 & 38.5 & 16.0 & \underline{35.0} & \underline{78.0} \\
            Eagle2.5 (w. Oracle) & 50.4 & 27.5 & 61.3 & 48.1 & 48.0 & \textbf{50.8} & \textbf{80.8} \\
            \bottomrule
            \end{tabular}%
            }
        \end{table}

    \vspace{-4mm}

    \subsection{Adaptive Reasoning: When to Think and When to Perceive}
    \label{sec:5.2}
        Video-Oasis emphasizes strict spatio-temporal dependencies that often require multi-hop understanding.
        This makes reasoning-depth control an important design axis: models must decide not only how to perceive the video, but also when deeper reasoning is necessary.
        To this end, we employ Qwen3-VL (8B)~\cite{qwem3_vl} as the base model and compare: (i) the instruction-following mode, (ii) the thinking mode, and (iii) adaptive thinking via VideoAuto-R1~\cite{videoautor1}.
        As detailed in \Cref{tab:main10}, adaptive thinking ($\text{Qwen3-VL}_\text{AutoR1}$) outperforms an always-on thinking strategy ($\text{Qwen3-VL}_\text{think.}$), suggesting that reasoning depth should be adjusted to the question rather than fixed in advance.
        
        \begin{table}[ht]
            \vspace{-4mm}
            \centering
            \caption{Ablation study of reasoning depth modulation. The metric is accuracy.}
            \label{tab:main10}
            \setlength{\tabcolsep}{4pt}
            \resizebox{0.9\textwidth}{!}{%
            \begin{tabular}{l c c c c c c}
            \toprule
            \textbf{Method} & \textbf{\small\makecell{Fine.\\Perception}} & \textbf{\small\makecell{Spatial\\World}} & \textbf{\small\makecell{Temporal\\Dynamics}} & \textbf{\small\makecell{Causal\\Logical}} & \textbf{\small\makecell{Global\\Narrative}} & 
            \textbf{Overall} \\
            \midrule
            $\text{Qwen3-VL}_\text{inst.}$~\cite{qwem3_vl}   & 27.0 & 42.4 & 36.5 & 28.0 & 21.5 & 33.8 \\
            $\text{Qwen3-VL}_\text{think.}$~\cite{qwem3_vl}  & 29.0 & 41.6 & 37.7 & 27.7 & 23.2 & 34.6 \\
            $\text{Qwen3-VL}_\text{AutoR1}$~\cite{videoautor1}   & 27.5 & 44.3 & 39.5 & 31.1 & 28.9 & 36.8 \\
            \midrule
            $\text{Qwen3-VL}_\text{voting}$ & \underline{38.4} & \textbf{57.7} & \underline{49.4} & \underline{37.8} & \underline{30.1} & \underline{46.2} \\
            Gemini-2.5-Pro~\cite{gemini25}   & \textbf{40.2} & \underline{49.8} & \textbf{50.9} & \textbf{45.4} & \textbf{43.0} & \textbf{46.7} \\
            \bottomrule
            \end{tabular}%
            }
            \vspace{-4mm}
        \end{table}

        We further ask how much performance could improve if the model chose the better mode for each question.
        To explore this, we introduce an oracle ensemble baseline ($\text{Qwen3-VL}_\text{voting}$), which considers a response correct if either the instruction-following ($\text{Qwen3-VL}_\text{inst.}$) or the thinking mode ($\text{Qwen3-VL}_\text{think.}$) successfully solves the task. 
        By simulating an optimal selection between thinking and non-thinking states, the voting baseline reaches 46.2, nearly closing the gap with the frontier-level Gemini-2.5-Pro (46.7).
        This finding yields a crucial insight: \textbf{the strategic optimization of when to think can be as impactful as the raw scale of the model's architecture.}

    \subsection{Training Paradigms}
    \label{sec:5.3}
        In this section, we review different training paradigms and conduct systematic experiments to compare their effectiveness for video understanding. 
        Specifically, our empirical analysis is twofold: (i) a direct performance comparison between supervised fine-tuning (SFT) and reinforcement learning with verifiable rewards (RLVR), and (ii) a targeted investigation into RLVR reward designs to determine the critical components for complex video understanding.
        Our empirical evaluation, leveraging Qwen2.5-VL~\cite{qwen25vl} as the unified base model in \Cref{tab:main11}, yields the following key insights:
        \begin{enumerate}
            \item \textbf{Effectiveness of long-context SFT:} Eagle2.5~\cite{eagle25} demonstrates a clear margin of improvement over the baseline Qwen2.5-VL~\cite{qwen25vl}, improving overall accuracy from 29.2\% to 34.5\% without relying on RLVR.
            This indicates that advanced long-context optimization can serve as a key factor in enhancing general spatio-temporal reasoning.
            \item \textbf{RLVR Reward Designs:} The comparative results among RLVR-based models and the SFT baseline reveal a non-linear performance trend: Video-R1~\cite{videor1} (26.3\%) $<$ Qwen2.5-VL~\cite{qwen25vl} (29.2\%) $<$ $\text{VideoAuto-R1}_\text{Qwen2.5}$~\cite{videoautor1} (32.7\%).
            These divergent outcomes indicate that reward formulation is critical to the success of RL-based video training.
            Despite its importance, the systematic investigation of reward structures for video reasoning remains largely underexplored, representing a potential area for future research.
            \item \textbf{Complementary Strengths of SFT and RLVR:} Our results suggest that SFT and RLVR offer complementary strengths rather than a single superior path. 
            While well-optimized SFT, as represented by Eagle2.5~\cite{eagle25}, improves overall accuracy, RLVR with grounding rewards, as represented by VideoAuto-R1~\cite{videoautor1}, provides larger gains on specific reasoning challenges such as \textit{Global Narrative} (21.2\%$\rightarrow$28.6\%).
            Recent studies~\cite{limitofrlvr, chu2025sft} similarly explore the complementary roles of SFT and RLVR, but remain largely confined to linguistic reasoning or static image domains, motivating their extension to video understanding.
        \end{enumerate}
        
        \begin{table}[ht]
            \vspace{-6mm}
            \centering
            \caption{Comparison of SFT and RLVR training paradigms for video understanding. All models share the same base LLM, Qwen2.5-VL~\cite{qwen25vl}. The metric is accuracy.}
            \label{tab:main11}
            \setlength{\tabcolsep}{4pt}
            \resizebox{1.0\textwidth}{!}{%
            \begin{tabular}{l c c c c c c c c}
            \toprule
            \multirow{2}{*}{\raisebox{0.5ex}{\textbf{Models}}} & \multicolumn{2}{c}{\textbf{\small Rewards}} & \multirow{2}{*}{\raisebox{0.5ex}{\textbf{\small\makecell{Fine.\\Perception}}}} & \multirow{2}{*}{\raisebox{0.5ex}{\textbf{\small\makecell{Spatial\\World}}}} & \multirow{2}{*}{\raisebox{0.5ex}{\textbf{\small\makecell{Temporal\\Dynamics}}}} & \multirow{2}{*}{\raisebox{0.5ex}{\textbf{\small\makecell{Causal\\Logical}}}} & \multirow{2}{*}{\raisebox{0.5ex}{\textbf{\small\makecell{Global\\Narrative}}}} & \multirow{2}{*}{\raisebox{0.5ex}{\textbf{Overall}}} \\
            \cmidrule(lr){2-3}
            & \textbf{\small QA} & \textbf{\small Grounding} & & & & & & \\
            \midrule
            Qwen2.5-VL~\cite{qwen25vl}    & - & - & 23.3 & 28.7 & 32.3 & 28.6 & 21.2 & 29.2 \\
            Eagle2.5~\cite{eagle25}      & - & - & \textbf{26.9} & \textbf{31.0} & \textbf{39.7} & \underline{33.2} & \underline{22.7} & \textbf{34.5} \\
            \midrule
            Video-R1~\cite{videor1}      & \textbf{$\checkmark$} & - & 24.0 & 24.0 & 29.1 & 27.3 & 18.4 & 26.3 \\
            $\text{VideoAuto-R1}_\text{Qwen2.5}$~\cite{videoautor1}  & \textbf{$\checkmark$} & \textbf{$\checkmark$} & \underline{25.4} & \underline{29.7} & \underline{35.9} & \textbf{33.7} & \textbf{28.6} & \underline{32.7} \\
            \bottomrule
            \end{tabular}%
            }
        \end{table}
    In summary, Video-Oasis not only diagnoses limitations in evaluation protocols but also reveals key insights into the algorithmic components of video understanding through comprehensive ablation studies. 
    These findings provide practical guidance for designing stronger video understanding methods.
}

\section{Conclusion}
{
    In this work, we establish Video-Oasis, a rigorous diagnostic suite for robust video understanding. 
    Through this diagnostic lens, we analyze vulnerabilities in existing benchmarks with respect to visual and temporal dependency and re-examine the current landscape of video understanding.
    Beyond diagnosis, our algorithmic exploration provides several insights, highlighting temporal grounding and adaptive reasoning as primary drivers of spatio-temporal reasoning. 
    We further identify the balance between SFT and RLVR as an open question for future Video-LLM training.
    Our work is fully reproducible, and we open-source the entire pipeline to enable large-scale auditing of existing datasets and provide an extensible evaluation protocol for the community.
    We hope Video-Oasis serves as a foundation for developing more rigorous benchmarks and drives the next generation of models toward robust video understanding.
    
    \noindent\textbf{Discussion.}
    Diagnostic tests such as caption-based or shuffled-frame evaluation may preserve partial temporal cues, leading to false positives or false negatives.
    To reduce this risk, Video-Oasis combines complementary diagnostic axes with cross-model consensus and human verification.
    We further view Video-Oasis as a reproducible and configurable audit suite that can be adapted to different evaluation goals by adjusting its tests.

}

\section*{Acknowledgements}
{
    This work was supported by the NAVER Cloud Corporation and partly supported by the National Research Foundation of Korea (NRF) grant funded by the Korea government (MSIT) (RS-2025-00553785, 40\%), the IITP under the virtual convergence support program to nurture the best talents grant funded by the Korea government (MSIT) (IITP-2026-RS-2023-00254529, 40\%), and the IITP under the Leading Generative AI Human Resources Development grant funded by the Korea government (MSIT) (IITP-2026-RS-2026-25544647, 20\%).
}




%
%
\bibliographystyle{splncs04}
\bibliography{ref}

\clearpage
\appendix
\setcounter{section}{0}
\renewcommand{\thesection}{\Alph{section}}
\renewcommand{\thesubsection}{\Alph{section}.\arabic{subsection}}
\renewcommand{\thesubsubsection}{\Alph{section}.\arabic{subsection}.\arabic{subsubsection}}

\setcounter{table}{0}
\renewcommand{\thetable}{S\arabic{table}}
\setcounter{figure}{0}
\renewcommand{\thefigure}{S\arabic{figure}}

\title{Video-Oasis: Rethinking Evaluation of \\ Video Understanding \\ (Supplementary Material)} 
\author{}
\institute{}
\titlerunning{Video-Oasis: Rethinking Evaluation of Video Understanding}
\authorrunning{G. Lim et al.}
\maketitle
\vspace{-6mm}

The supplementary material is organized as follows:

\begin{itemize}[topsep=2pt,itemsep=1pt]
\item \Cref{sec:a}: robustness analysis of Video-Oasis under alternative diagnostic model configurations.
\item \Cref{sec:b}: benchmark-wise evaluation results, including shortcut filtering outcomes and diagnostic test results across benchmarks.
\item \Cref{sec:c}: implementation details of Video-Oasis.
\item \Cref{sec:d}: statistics, qualitative examples, and annotation prompts for the distilled video-native challenges.
\item \Cref{sec:e}: additional experimental results related to reproducibility.

\end{itemize} \vspace{-4mm}

\section{Robustness of Video-Oasis} \label{sec:a}

    We perform an ablation study with alternative models in the diagnostic suite.
    For the visual dependency tests (Blind, Audio, and Summary), we replace the Video-LLMs with language-only LLMs~\cite{llama3,mistral,qwen3}, and for the Center-Frame and Frame Shuffling tests, we use different backbones (InternVL-3.5~\cite{internvl35}, Video-R1~\cite{videor1}, VideoLLaMA3~\cite{videollama}, MiMo-VL-SFT~\cite{mimo}, and LLaVA-Video~\cite{zhang2024llava}). 
    \Cref{tab:multi_voting_6_tables} reports the diagnostic test results under these alternative model configurations.
    Following the main paper, we define a shortcut sample only when all three models correctly answer the question.
    As shown in \Cref{tab:robustness_overlap}, the resulting shortcut sets remain highly consistent with those obtained under the main diagnostic model configuration.
    All overlaps exceed 87\%, indicating robustness to model selection.
            
    \begin{table}[h]
        \centering
        \footnotesize 
        \setlength{\tabcolsep}{5pt} 
        \begin{subtable}[t]{0.31\textwidth}
            \centering
            \begin{tabular}{lc}
                \toprule
                \rowcolor[gray]{0.9} \textbf{LLM} & \textbf{Acc.} \\
                \midrule
                Mistral & 29.3 \\
                Qwen3 & 33.1 \\
                Llama 3.1 & 31.6 \\
                \bottomrule
            \end{tabular}
            \caption{Blind Test}
        \end{subtable}
        \hfill
        \begin{subtable}[t]{0.31\textwidth}
            \centering
            \begin{tabular}{lc}
                \toprule
                \rowcolor[gray]{0.9} \textbf{LLM} & \textbf{Acc.} \\
                \midrule
                Mistral & 37.5 \\
                Qwen3 & 39.7 \\
                Llama 3.1 & 41.2 \\
                \bottomrule
            \end{tabular}
            \caption{Audio}
        \end{subtable}
        \hfill
        \begin{subtable}[t]{0.31\textwidth}
            \centering
            \begin{tabular}{lc}
                \toprule
                \rowcolor[gray]{0.9} \textbf{LLM} & \textbf{Acc.} \\
                \midrule
                Mistral & 38.5 \\
                Qwen3 & 42.6 \\
                Llama 3.1 & 40.5 \\
                \bottomrule
            \end{tabular}
            \caption{Summary}
        \end{subtable}
    
        \vspace{0.2em} 
    

        \begin{subtable}[t]{0.31\textwidth}
            \centering
            \begin{tabular}{lc}
                \toprule
                \rowcolor[gray]{0.9} \textbf{VLM} & \textbf{Acc.} \\
                \midrule
                $\text{CLIP}$ & 31.4 \\
                $\text{Long-CLIP}$ & 32.8 \\
                $\text{EVA-CLIP}$ & 32.7 \\
                \bottomrule
            \end{tabular}
            \caption{Bag-of-Frames}
        \end{subtable}
        \hfill
        \begin{subtable}[t]{0.31\textwidth}
            \centering
            \begin{tabular}{lc}
                \toprule
                \rowcolor[gray]{0.9} \textbf{Video-LLM} & \textbf{Acc.} \\
                \midrule
                Eagle2.5 & 42.2 \\
                Video-R1 & 38.6 \\
                InternVL-3.5 & 39.2 \\
                MiMo-VL-SFT & 38.6 \\
                LLaVA-Video & 39.2 \\
                VideoLLaMA3 & 42.2 \\

                \bottomrule
            \end{tabular}
            \caption{Center-Frame}
        \end{subtable}
        \hfill
        \begin{subtable}[t]{0.31\textwidth}
            \centering
            \begin{tabular}{lc}
                \toprule
                \rowcolor[gray]{0.9} \textbf{Video-LLM} & \textbf{Acc.} \\
                \midrule
                Eagle2.5 & 52.2 \\
                Video-R1 & 46.3 \\
                InternVL-3.5 & 49.8 \\
                MiMo-VL-SFT & 46.1 \\
                LLaVA-Video & 47.4 \\
                VideoLLaMA3 & 48.4 \\
                \bottomrule
            \end{tabular}
            \caption{Frame Shuffling}
        \end{subtable}
        \caption{Quantitative results of Video-Oasis under different diagnostic models.} \vspace{-12mm}
        \label{tab:multi_voting_6_tables}
    \end{table}

    \begin{table}[h]
    \centering
    \scriptsize
    \caption{Overlap of shortcut sets under alternative diagnostic model combinations.}
    \label{tab:robustness_overlap}
    \resizebox{0.95\columnwidth}{!}{%
    \begin{tabular}{l l l | c}
    \toprule
    \textbf{Model 1} & \textbf{Model 2} & \textbf{Model 3} & \textbf{Overlap} \\
    \midrule
    Eagle2.5 (8B) & InternVL-3.5 (8B) & Video-R1 (8B) & 90.0\% \\
    InternVL-3.5 (8B) & VideoLLaMA3 (8B) & MiMo-VL-SFT (8B) & 88.6\% \\
    InternVL-3.5 (8B) & VideoLLaMA3 (8B) & LLaVA-Video (8B) & 87.5\% \\
    VideoLLaMA3 (8B) & MiMo-VL-SFT (8B) & LLaVA-Video (8B) & 87.8\% \\
    \bottomrule
    \end{tabular}%
    }
    \end{table}
    
\section{Benchmark-wise Evaluation} \label{sec:b}

    This section presents additional benchmark-wise analyses that were not included in the main paper. 
    \Cref{sec:b.1} reports the proportion of shortcut samples identified for each benchmark, and \Cref{sec:b.2} presents the diagnostic test results.
    
    \subsection{Filtering Results of Video-Oasis} \label{sec:b.1}

        \Cref{tab:supp_benchmark_stats} summarizes the per-benchmark filtering results. 
        For each benchmark, we report the number of QA samples before filtering (original) and after applying Video-Oasis (remaining), where the filtering ratio denotes the proportion of removed samples. 
        We also evaluate Qwen2.5-VL (7B)~\cite{qwen25vl} on both sets and report the absolute performance gap.
        The relationship between the filtering ratio and the performance gap is not uniform across benchmarks. 
        If Video-Oasis simply selected difficult questions, benchmarks with higher original accuracy would be expected to exhibit both higher filtering ratios and larger performance gaps, while those with lower original accuracy would show the opposite trend.
        However, the results do not follow such a monotonic trend. 
        For example, MVBench~\cite{mvbench} achieves a higher original accuracy (71.2) than EgoSchema~\cite{egoschema} (62.4), yet EgoSchema exhibits both a higher filtering ratio (76.4 vs. 66.9) and a substantially larger performance gap (40.5 vs. 20.7). If Video-Oasis merely selected difficult questions, the opposite pattern would be expected.
        These observations suggest that Video-Oasis does not simply collect difficult questions but instead identifies shortcut-prone problems that fail to enforce video understanding. 
        \vspace{-4mm}

        \begin{table}[h]
        \centering
        \caption{Per-benchmark statistics before and after Video-Oasis filtering.}
        \label{tab:supp_benchmark_stats}
        
        \setlength{\tabcolsep}{4pt}
        \renewcommand{\arraystretch}{1.05}
        
        \resizebox{\linewidth}{!}{
        \begin{tabular}{llrr}
        \toprule
        \multirow{3}{*}{\textbf{Group}} &
        \multirow{3}{*}{\textbf{Benchmark}} &
        \multicolumn{1}{c}{\textbf{Question Samples}} &
        \multicolumn{1}{c}{\textbf{Accuracy (\%) on}} \\
         & & \multicolumn{1}{c}{\textbf{remaining/original}} &
             \multicolumn{1}{c}{\textbf{remaining/original}} \\
         & & \multicolumn{1}{c}{(filtering ratio)} &
             \multicolumn{1}{c}{(performance gap)} \\
        \midrule
        \textbf{Spatial}
        & EgoSchema~\cite{egoschema}           & 118/500 (76.4\%)     & 21.9/62.4 (40.5\%) \\
        & ImplicitQA~\cite{implicitqa}         & 356/766 (53.5\%)     & 24.5/48.6 (24.1\%) \\
        & VSI-Bench~\cite{vsibench}            & 1,550/2,490 (37.8\%) & 30.3/37.7 (7.40\%) \\
        \midrule
        \textbf{Temporal}
        & TVBench~\cite{tvbench}               & 1,174/2,205 (46.8\%) & 38.3/50.4 (12.1\%) \\
        & VCR-Bench~\cite{vcrbench}            & 255/511 (50.1\%)     & 34.0/53.8 (19.8\%) \\
        & RTV-Bench~\cite{rtvbench}            & 1,920/4,608 (58.3\%) & 22.5/45.5 (23.0\%) \\
        \midrule
        \textbf{Reasoning}
        & Video-Holmes~\cite{videoholmes}      & 958/1,837 (47.8\%)   & 24.1/43.6 (19.5\%) \\
        & MINERVA~\cite{minerva}               & 901/1,358 (33.7\%)   & 22.9/35.7 (12.8\%) \\
        & MMR-V~\cite{mmrv}                    & 658/1,257 (47.7\%)   & 17.8/45.9 (28.1\%) \\
        \midrule
        \textbf{General}
        & VideoMME~\cite{videomme}            & 633/2,700 (76.6\%)   & 32.5/65.3 (32.8\%) \\
        & MVBench~\cite{mvbench}               & 994/3,000 (66.9\%) & 50.5/71.2 (20.7\%) \\
        & LVBench~\cite{lvbench}               & 740/1,345 (45.0\%)   & 24.3/41.7 (17.4\%) \\
        & LongVideoBench~\cite{longvideobench} & 545/1,337 (59.2\%)   & 30.4/59.7 (29.3\%) \\
        & MLVU~\cite{mlvu}                     & 231/502 (54.0\%)     & 27.3/53.6 (26.3\%) \\
        \midrule
        \rowcolor{gray!10}
        \textbf{Total} & & \textbf{11,033/24,416 (54.8\%)} & \textbf{29.2/51.2 (22.0\%)} \\
        \bottomrule
        \end{tabular}}
        \vspace{-8mm}
        \end{table}

    \subsection{Diagnostic Test Results} \label{sec:b.2}
        We report benchmark-wise results for each diagnostic test to provide a detailed breakdown across datasets. 
        The results are summarized in \Cref{tab:blind,tab:audio,tab:narrative,tab:center,tab:shuffling,tab:bof}, which report the model accuracy under the proposed diagnostic configurations.
        For the Audio Test, as shown in \Cref{tab:audio}, certain benchmarks are excluded from evaluation because their videos do not contain encoded audio tracks, the audio tracks are unavailable on YouTube, or the videos contain no speech. 
        Consequently, we evaluate only the subset of samples for which audio-based evaluation is feasible.    
                
        \begin{table}[h]
        \centering
        \caption{Benchmark-wise results (\%) for the Blind Test, where the model answers without visual input or auxiliary context, relying only on linguistic priors.}
        \footnotesize
        \setlength{\tabcolsep}{4pt}
        \begin{tabular}{lcccc}
        \toprule
        \multirow{2}{*}{\textbf{Benchmark}} & \multicolumn{4}{c}{\scriptsize \textbf{Blind}} \\
        \cmidrule(lr){2-5}
         & {\scriptsize \textbf{Eagle2.5~\cite{eagle25}}} & {\scriptsize \textbf{Qwen2.5-VL}~\cite{qwen25vl}} & {\scriptsize \textbf{Qwen3-VL}~\cite{qwem3_vl}} & {\scriptsize \textbf{Random}} \\
        \midrule
        Egoschema~\cite{egoschema} & 30.6 & 29.1 & 35.0 & 20.0 \\
        ImplicitQA~\cite{implicitqa} & 39.0 & 38.1 & 40.3 & 29.3 \\
        MINERVA~\cite{minerva} & 24.2 & 20.0 & 23.6 & 20.0 \\
        RTV-Bench~\cite{rtvbench} & 37.1 & 34.6 & 35.5 & 30.4 \\
        VSI-Bench~\cite{vsibench} & 25.0 & 29.6 & 31.5 & 28.7 \\
        MVBench~\cite{mvbench} & 37.5 & 42.6 & 38.9 & 29.7 \\
        LongVideoBench~\cite{longvideobench} & 46.2 & 42.1 & 43.6 & 21.3 \\
        LVBench~\cite{lvbench} & 28.5 & 23.5 & 26.8 & 25.0 \\
        MLVU~\cite{mlvu} & 24.4 & 27.3 & 25.6 & 16.7 \\
        MMR-V~\cite{mmrv} & 35.7 & 30.3 & 34.4 & 9.6 \\
        TVBench~\cite{tvbench} & 37.0 & 34.7 & 40.9 & 33.7 \\
        VCR-Bench~\cite{vcrbench} & 42.0 & 37.8 & 39.7 & 24.1 \\
        Video-Holmes~\cite{videoholmes} & 36.6 & 32.5 & 35.9 & 16.7 \\
        VideoMME~\cite{videomme} & 43.4 & 38.2 & 43.9 & 25.0 \\
        \midrule
        \textbf{Total} & \textbf{35.6} & \textbf{33.5} & \textbf{36.2} & \textbf{25.6} \\
        \bottomrule
        \end{tabular}
        \label{tab:blind}
        \end{table}

        \begin{table}[h]
        \centering
        \caption{Benchmark-wise results (\%) for the Audio Test, where the model answers using only the speech transcript from the video's audio.}
        \footnotesize
        \setlength{\tabcolsep}{4pt}
        \begin{tabular}{lcccc}
        \toprule
        \multirow{2}{*}{\textbf{Benchmark}} & \multicolumn{4}{c}{\scriptsize \textbf{Audio}} \\
        \cmidrule(lr){2-5}
         & {\scriptsize \textbf{Eagle2.5~\cite{eagle25}}} & {\scriptsize \textbf{Qwen2.5-VL}~\cite{qwen25vl}} & {\scriptsize \textbf{Qwen3-VL}~\cite{qwem3_vl}} & {\scriptsize \textbf{Random}} \\
        \midrule
        EgoSchema~\cite{egoschema} & - & - & - & - \\
        ImplicitQA~\cite{implicitqa} & - & - & - & - \\
        MINERVA~\cite{minerva} & - & - & - & - \\
        RTV-Bench~\cite{rtvbench} & 37.5 & 37.3 & 37.0 & 30.2 \\
        VSI-Bench~\cite{vsibench} & - & - & - & - \\
        MVBench~\cite{mvbench} & 40.3 & 38.0 & 34.5 & 28.9 \\
        LongVideoBench~\cite{longvideobench} & 50.5 & 54.3 & 45.7 & 20.8 \\
        LVBench~\cite{lvbench} & - & - & - & - \\
        MLVU~\cite{mlvu} & 42.5 & 52.1 & 46.2 & 16.7 \\
        MMR-V~\cite{mmrv} & 32.0 & 31.0 & 35.6 & 9.5 \\
        TVBench~\cite{tvbench} & 43.3 & 37.4 & 38.6 & 35.7 \\
        VCR-Bench~\cite{vcrbench} & 48.6 & 46.5 & 42.7 & 25.0 \\
        Video-Holmes~\cite{videoholmes} & 37.8 & 38.2 & 39.9 & 16.7 \\
        VideoMME~\cite{videomme} & 65.4 & 64.0 & 64.5 & 25.0 \\
        \midrule
        \textbf{Total} & \textbf{47.6} & \textbf{46.8} & \textbf{45.9} & \textbf{24.9} \\
        \bottomrule
        \end{tabular}
        \label{tab:audio}
        \end{table}

        \begin{table}[h]
        \centering
        \caption{Benchmark-wise results (\%) for the Summary Test, where the model answers using concatenated video captions as textual context.}
        \footnotesize
        \setlength{\tabcolsep}{4pt}
        \begin{tabular}{lcccc}
        \toprule
        \multirow{2}{*}{\textbf{Benchmark}} & \multicolumn{4}{c}{\scriptsize \textbf{Summary}} \\
        \cmidrule(lr){2-5}
         & {\scriptsize \textbf{Eagle2.5~\cite{eagle25}}} & {\scriptsize \textbf{Qwen2.5-VL}~\cite{qwen25vl}} & {\scriptsize \textbf{Qwen3-VL}~\cite{qwem3_vl}} & {\scriptsize \textbf{Random}} \\
        \midrule
        EgoSchema~\cite{egoschema} & 69.8 & 65.9 & 70.5 & 20.0 \\
        ImplicitQA~\cite{implicitqa} & 41.8 & 40.6 & 40.3 & 29.3 \\
        MINERVA~\cite{minerva} & 27.8 & 25.6 & 27.9 & 20.0 \\
        RTV-Bench~\cite{rtvbench} & 41.9 & 41.0 & 42.2 & 30.4 \\
        VSI-Bench~\cite{vsibench} & 33.1 & 37.1 & 41.3 & 28.7 \\
        MVBench~\cite{mvbench} & 55.7 & 54.5 & 54.6 & 29.7 \\
        LongVideoBench~\cite{longvideobench} & 49.6 & 48.4 & 50.4 & 21.3 \\
        LVBench~\cite{lvbench} & 35.1 & 32.8 & 33.4 & 25.0 \\
        MLVU~\cite{mlvu} & 46.0 & 42.7 & 47.5 & 16.7 \\
        MMR-V~\cite{mmrv} & 39.1 & 34.6 & 38.2 & 9.6 \\
        TVBench~\cite{tvbench} & 43.9 & 42.8 & 42.4 & 33.7 \\
        VCR-Bench~\cite{vcrbench} & 45.7 & 46.6 & 50.2 & 24.1 \\
        Video-Holmes~\cite{videoholmes} & 39.3 & 35.5 & 41.0 & 16.7 \\
        VideoMME~\cite{videomme} & 54.8 & 52.4 & 57.6 & 25.0 \\
        \midrule
        \textbf{Total} & \textbf{44.0} & \textbf{42.6} & \textbf{45.0} & \textbf{25.6} \\
        \bottomrule
        \end{tabular}
        \label{tab:narrative}
        \end{table}

        \begin{table}[h]
        \centering
        \caption{Benchmark-wise results (\%) for the Center-Frame Test, where the model answers using only the center frame of the video.}
        \footnotesize
        \setlength{\tabcolsep}{4pt}
        \begin{tabular}{lcccc}
        \toprule
        \multirow{2}{*}{\textbf{Benchmark}} & \multicolumn{4}{c}{\scriptsize \textbf{Center-Frame}} \\
        \cmidrule(lr){2-5}
         & {\scriptsize \textbf{Eagle2.5~\cite{eagle25}}} & {\scriptsize \textbf{Qwen3-VL}~\cite{qwem3_vl}} & {\scriptsize \textbf{VideoAuto-R1}~\cite{videoautor1}} & {\scriptsize \textbf{Random}} \\
        \midrule
        EgoSchema~\cite{egoschema} & 49.8 & 58.1 & 58.4 & 20.0 \\
        ImplicitQA~\cite{implicitqa} & 39.8 & 38.0 & 41.8 & 29.3 \\
        MINERVA~\cite{minerva} & 27.3 & 23.3 & 25.2 & 20.0 \\
        RTV-Bench~\cite{rtvbench} & 40.7 & 40.8 & 41.7 & 30.4 \\
        VSI-Bench~\cite{vsibench} & 29.6 & 34.6 & 35.3 & 28.7 \\
        MVBench~\cite{mvbench} & 54.3 & 50.9 & 54.4 & 30.3 \\
        LongVideoBench~\cite{longvideobench} & 49.1 & 43.5 & 48.2 & 21.3 \\
        LVBench~\cite{lvbench} & 35.9 & 30.4 & 31.2 & 25.0 \\
        MLVU~\cite{mlvu} & 38.3 & 33.1 & 36.7 & 16.7 \\
        MMR-V~\cite{mmrv} & 47.6 & 41.0 & 46.0 & 9.6 \\
        TVBench~\cite{tvbench} & 39.9 & 38.5 & 43.4 & 33.4 \\
        VCR-Bench~\cite{vcrbench} & 40.9 & 37.6 & 40.9 & 24.1 \\
        Video-Holmes~\cite{videoholmes} & 42.0 & 35.9 & 42.3 & 16.7 \\
        VideoMME~\cite{videomme} & 50.7 & 47.5 & 50.4 & 25.0 \\
        \midrule
        \textbf{Total} & \textbf{42.2} & \textbf{40.2} & \textbf{43.0} & \textbf{25.6} \\
        \bottomrule
        \end{tabular}
        \label{tab:center}
        \end{table}

        \begin{table}[h]
        \centering
        \caption{Benchmark-wise results (\%) for the Frame Shuffling Test, where the temporal order of video frames is randomly permuted.}
        \footnotesize
        \setlength{\tabcolsep}{4pt}
        \begin{tabular}{lcccc}
        \toprule
        \multirow{2}{*}{\textbf{Benchmark}} & \multicolumn{4}{c}{\scriptsize \textbf{Frame Shuffling}} \\
        \cmidrule(lr){2-5}
         & {\scriptsize \textbf{Eagle2.5~\cite{eagle25}}} & {\scriptsize \textbf{Qwen3-VL}~\cite{qwem3_vl}} & {\scriptsize \textbf{VideoAuto-R1}~\cite{videoautor1}} & {\scriptsize \textbf{Random}} \\
        \midrule
        EgoSchema~\cite{egoschema} & 69.4 & 70.4 & 70.8 & 20.0 \\
        ImplicitQA~\cite{implicitqa} & 50.0 & 44.1 & 46.7 & 29.3 \\
        MINERVA~\cite{minerva} & 39.2 & 35.4 & 36.2 & 20.0 \\
        RTV-Bench~\cite{rtvbench} & 46.8 & 47.2 & 47.1 & 30.4 \\
        VSI-Bench~\cite{vsibench} & 35.5 & 46.4 & 45.8 & 28.7 \\
        MVBench~\cite{mvbench} & 69.1 & 64.7 & 66.2 & 30.3 \\
        LongVideoBench~\cite{longvideobench} & 63.7 & 59.4 & 60.6 & 21.3 \\
        LVBench~\cite{lvbench} & 49.4 & 43.4 & 42.8 & 25.0 \\
        MLVU~\cite{mlvu} & 54.6 & 52.6 & 51.8 & 16.7 \\
        MMR-V~\cite{mmrv} & 55.1 & 45.1 & 55.1 & 9.6 \\
        TVBench~\cite{tvbench} & 42.6 & 40.8 & 45.9 & 33.4 \\
        VCR-Bench~\cite{vcrbench} & 51.7 & 50.5 & 52.8 & 24.1 \\
        Video-Holmes~\cite{videoholmes} & 46.5 & 44.4 & 48.5 & 16.7 \\
        VideoMME~\cite{videomme} & 67.5 & 65.1 & 65.9 & 25.0 \\
        \midrule
        \textbf{Total} & \textbf{52.2} & \textbf{50.7} & \textbf{52.4} & \textbf{25.6} \\
        \bottomrule
        \end{tabular}
        \label{tab:shuffling}
        \end{table}

        \begin{table}[h]
        \centering
        \caption{Benchmark-wise results (\%) for the Bag-of-Frames Test, where frames are processed independently without modeling temporal relations.}
        \footnotesize
        \setlength{\tabcolsep}{4pt}
        \begin{tabular}{lcccc}
        \toprule
        \multirow{2}{*}{\textbf{Benchmark}} & \multicolumn{4}{c}{\scriptsize \textbf{Bag-of-Frames}} \\
        \cmidrule(lr){2-5}
         & {\scriptsize \textbf{CLIP~\cite{clip}}} & {\scriptsize \textbf{Long-CLIP}~\cite{longclip}} & {\scriptsize \textbf{EVA-CLIP}~\cite{evaclip}} & {\scriptsize \textbf{Random}} \\
        \midrule
        EgoSchema~\cite{egoschema} & 39.6 & 52.4 & 36.2 & 20.0 \\
        ImplicitQA~\cite{implicitqa} & 30.3 & 31.7 & 30.0 & 29.3 \\
        MINERVA~\cite{minerva} & 20.3 & 22.2 & 21.1 & 20.0 \\
        RTV-Bench~\cite{rtvbench} & 33.4 & 35.9 & 37.3 & 30.4 \\
        VSI-Bench~\cite{vsibench} & 33.1 & 28.3 & 34.3 & 28.7 \\
        MVBench~\cite{mvbench} & 41.0 & 42.2 & 41.4 & 30.3 \\
        LongVideoBench~\cite{longvideobench} & 31.3 & 32.2 & 34.3 & 21.3 \\
        LVBench~\cite{lvbench} & 30.0 & 31.5 & 29.4 & 25.0 \\
        MLVU~\cite{mlvu} & 29.7 & 33.7 & 27.3 & 16.7 \\
        MMR-V~\cite{mmrv} & 18.4 & 19.1 & 19.6 & 9.6 \\
        TVBench~\cite{tvbench} & 33.3 & 37.4 & 35.4 & 33.4 \\
        VCR-Bench~\cite{vcrbench} & 29.6 & 31.0 & 26.9 & 24.1 \\
        Video-Holmes~\cite{videoholmes} & 20.2 & 20.4 & 20.3 & 16.7 \\
        VideoMME~\cite{videomme} & 33.8 & 34.9 & 35.3 & 25.0 \\
        \midrule
        \textbf{Total} & \textbf{31.4} & \textbf{32.7} & \textbf{32.8} & \textbf{25.6} \\
        \bottomrule
        \end{tabular}
        \label{tab:bof}
        \end{table}
    \clearpage

\section{Implementation Details of Video-Oasis} \label{sec:c}

    \subsection{Visual Dependency Tests}
        In this setting, visual input is removed, and the model ($\mathcal{M}$) is evaluated based solely on textual context. 
        \begin{equation}
            \text{Response} = \mathcal{M}\left( \mathcal{V}, \mathcal{P}_{diag}(\mathcal{C}_{type}, \mathcal{Q}, \mathcal{O}) \right),
        \end{equation}
        
        \noindent where $\mathcal{V}$ denotes the visual input and $\mathcal{P}_{diag}$ is a system prompt (\Cref{tab:prompt}) that formats the context $\mathcal{C}_{type}$ together with the question $\mathcal{Q}$ and options $\mathcal{O}$. For all tests, we set $\mathcal{V} = \emptyset$ to remove the influence of visual context.
        
        \paragraph{\textbf{Blind Test ($\mathcal{C}_{blind} = \emptyset$)}}
        In this configuration, all auxiliary inputs are removed. 
        The model must rely solely on linguistic priors and internal world knowledge.
        
        \paragraph{\textbf{Audio Test ($\mathcal{C}_{audio} = \mathcal{T}_{whisper}$)}} 
        Here, the context is replaced by a text transcript $\mathcal{T}$ generated from the video's audio track. 
        We use Whisper-v3 (large)~\cite{whisper} to extract speech. 
        This setup identifies questions that can be solved solely from the speech transcript, indicating that they primarily require textual understanding rather than visual reasoning.
        
        \paragraph{\textbf{Summary Test ($\mathcal{C}_{summary} = \mathcal{S}_{concat}$)}} 
        The video is uniformly partitioned into 8 temporal chunks (16 frames each), where a single caption is extracted per chunk via CARE~\cite{care} and concatenated chronologically to form $\mathcal{S}_{concat}$. 
        Since our setup relies on simple caption concatenation without sophisticated summarization, success in this test suggests that the task may rely more on text-based reasoning, such as pattern matching or attribute recognition, than on grounded visual perception from the raw video.

        \begin{table}[h!] \centering \begin{tcolorbox}[colback=gray!5, colframe=gray!50, arc=2mm, boxrule=0.5pt, left=2mm, right=2mm, top=2mm, bottom=2mm] \footnotesize \textbf{System Prompt:} \\ You are a helpful assistant. Select the best answer to the following multiple-choice question based on the provided context and options. \\ \textbf{Question ($\mathcal{Q}$):} \{question\} \\ \textbf{Context ($\mathcal{C}_{type}$):} \textit{\{Empty / Audio Transcript / Summary\}} \\ \textbf{Options ($\mathcal{O}$):} \{options\} \\ \textbf{Output Constraint:} \\ Respond with only the letter (A, B, C, D...) of the correct option. \end{tcolorbox} \caption{Prompt template for visual dependency tests with different contexts.} \label{tab:prompt} \vspace{-8mm} \end{table}
        
    \subsection{Temporal Dependency Tests}
    
        In this setting, the intrinsic temporal structure of the video is disrupted by varying both the frame sampling strategy and the model perspective.
        \begin{equation}
            \text{Response} = \mathcal{M}\left( \mathcal{S}(\mathcal{V}), \mathcal{P}_{diag}(\mathcal{Q}, \mathcal{O}) \right),
        \end{equation}
        
        \noindent where $\mathcal{S}(\mathcal{V})$ denotes a subset of the video and $\mathcal{M}$ denotes the model.
    
        \paragraph{\textbf{Temporal Context Strategy ($\mathcal{S}$)}}
        We vary the frame selection strategy $\mathcal{S}$ to examine how models respond under different conditions.
        \begin{itemize}
            \item \textbf{Center Frame ($\mathcal{S}_{\text{center}}$):} Returns a single frame from the temporal center of $\mathcal{V}$. 
            This test identifies whether the task can be solved using spatial cues alone, often due to redundancy in video frames.
            
            \item \textbf{Frame Shuffling ($\mathcal{S}_{\text{shuffle}}$):} Returns a randomly permuted subset of 128 uniformly sampled frames from $\mathcal{V}$. 
            This process disrupts the chronological order of the video.
            
            \item \textbf{Top-$k$ Matching ($\mathcal{S}_{\text{topk}}$):} Returns a subset of $k$ frames ($k=32$) exhibiting the highest cosine similarity with the query $\mathcal{Q}$ in the VLM embedding space.
        \end{itemize}
    
        \paragraph{\textbf{Model Perspective ($\mathcal{M}$)}}
        We categorize the models into two types based on how they process the configured temporal subset $\mathcal{S}(\mathcal{V})$:
        
        \begin{itemize}
            \item \textbf{MLLM ($\mathcal{M}_{\text{MLLM}}$):} Multimodal large language models~\cite{eagle25,qwem3_vl,videoautor1}, pretrained to reason over temporal dynamics in video data. 
            We evaluate the model responses under temporal disruption (\eg, center-frame ($\mathcal{S}_{\text{center}}$) or shuffling ($\mathcal{S}_{\text{shuffle}}$)) to assess whether the question relies on temporal context.
        
            \item \textbf{VLM ($\mathcal{M}_{\text{VLM}}$):} Vision-language models pretrained on static image-text pairs (\textit{e.g.}, CLIP~\cite{clip,evaclip,longclip}), which process frames independently. 
            We evaluate them via top-$k$ cosine-similarity matching ($\mathcal{S}_{\text{topk}}$) between frame embeddings and answer candidates. This setup is temporal-agnostic, as the model processes frames independently without modeling temporal relations.
        \end{itemize}
    
        These tests diagnose whether the question depends on temporal context by systematically disrupting the temporal structure of the input video. The overall diagnostic procedure for the temporal dependency tests is summarized in \Cref{alg:diagnostic}.
    
        \begin{algorithm}[t]
        \caption{Temporal Dependency Diagnostic Procedure}
        \label{alg:diagnostic}
        \begin{algorithmic}[1]
        \State \textbf{Input:} Video $\mathcal{V}$, Question $\mathcal{Q}$, Options $\mathcal{O}$, Cosine similarity function $\text{sim}(\cdot, \cdot)$
        \State \textbf{Output:} Set of Diagnostic Responses $\{R_{\mathcal{S}}\}$
        
        \For{each sampling strategy $\mathcal{S} \in \{\mathcal{S}_{\text{center}}, \mathcal{S}_{\text{shuffle}}, \mathcal{S}_{\text{topk}}\}$}
            \If{$\mathcal{S} = \mathcal{S}_{\text{topk}}$}
                \State $\mathcal{V}_{\text{subset}} \gets \text{Top-K}(\mathcal{V}, \mathcal{Q})$ \Comment{retrieve frames most similar to $\mathcal{Q}$}
                \State $R_{\mathcal{S}} \gets \arg\max_{o \in \mathcal{O}} \sum_{f \in \mathcal{V}_{\text{subset}}} \text{sim}(\mathcal{M}_{\text{VLM}}(f), \mathcal{M}_{\text{VLM}}(o))$ \Comment{BoF Matching}
            \Else
                \State $\mathcal{V}_{\text{subset}} \gets \mathcal{S}(\mathcal{V})$ \Comment{Center-frame or random shuffling}
                \State $R_{\mathcal{S}} \gets \mathcal{M}_{\text{MLLM}}(\mathcal{V}_{\text{subset}}, \mathcal{P}_{diag}(\mathcal{Q}, \mathcal{O}))$ \Comment{Temporal disruption test}
            \EndIf
        \EndFor
        
        \State \Return $\mathcal{R} = \{R_{\mathcal{S}_{\text{center}}}, R_{\mathcal{S}_{\text{shuffle}}}, R_{\mathcal{S}_{\text{topk}}}\}$
        \end{algorithmic}
        \end{algorithm} 

    \clearpage

    \subsection{Ambiguity Tests}
    
        In this setting, we introduce diagnostic tests to identify potential annotation issues arising during dataset construction:
        \begin{itemize}
            \item \textbf{Consistency Test.} We employ five different models~\cite{eagle25,qwem3_vl,internvl35,videoautor1,videor1} and identify samples where all models produce different predictions. 
            Such strong disagreement suggests that the question may admit multiple plausible interpretations, indicating potential annotation ambiguity.
        
            \item \textbf{Redundancy Test.} Using Eagle2.5~\cite{eagle25}, we divide each video into 8 temporal chunks (16 frames each) and evaluate the model on each chunk independently. If all chunks lead to the correct answer, the question may not rely on specific temporal evidence and can be answered from multiple segments, indicating that the annotation may be weakly constrained.

            \item \textbf{Sensitivity Test.} We manually inspect samples that are answered correctly even after frame shuffling. If the question still requires temporal ordering despite the shuffled-input success, the sample is restored to avoid false positives from the temporal perturbation test.
            
        \end{itemize}

\section{Video-Native Challenge: Statistics and Examples} \label{sec:d}
    
    \subsection{Video-Native Challenges Statistics} \label{sec:d.1}
    
        \Cref{tab:statistics}(a) presents the distribution of the identified video-native challenge categories. 
        Temporal dynamics and tracking constitute the largest portion (51.0\%) of the identified challenges, while the remaining categories also contain sufficient samples to enable comprehensive evaluation. 
        \Cref{tab:statistics}(b) shows the distribution of video durations. The suite spans a broad range from short clips under 15 seconds (19.5\%) to long videos exceeding 10 minutes (18.9\%), confirming that Video-Oasis preserves diversity in temporal scale.
        
        \begin{table}[h]
            \centering
            \caption{Statistics of the identified video-native challenges: (a) category distribution and (b) video duration statistics.}
            \label{tab:statistics}
            \begin{minipage}[t]{0.55\linewidth}
            \centering
            \small
            \vspace{0pt}
            (a) Challenge Categories \\[4pt]
            \resizebox{\linewidth}{!}{%
            \begin{tabular}{l r r}
            \toprule
            \textbf{Category} & \textbf{\# Samples} & \textbf{Ratio} \\
            \midrule
            Fine-Grained Perception        & 1,107 & 10.0 \\
            Spatial World Understanding    & 1,789 & 16.2 \\
            Temporal Dynamics \& Tracking  & 5,631 & 51.0 \\
            Causality \& Logical Reasoning & 1,414 & 12.8 \\
            Global Narrative               & 1,092 &  9.9 \\
            \midrule
            \rowcolor{gray!10}
            \textbf{Total}       & \textbf{11,033} & \\
            \bottomrule
            \end{tabular}}
            \end{minipage}
            \hfill
            \begin{minipage}[t]{0.35\linewidth}
            \centering
            \small
            \vspace{0pt}
            (b) Video Duration \\[4pt]
            \resizebox{\linewidth}{!}{%
            \begin{tabular}{l r r}
            \toprule
            \textbf{Duration} & \textbf{\# Videos} & \textbf{Ratio} \\
            \midrule
            $\sim$15s            & 969 & 19.6 \\
            $\sim$1min           & 1,160 & 23.5 \\
            $\sim$10min          & 1,887 & 38.2 \\
            10min+               & 922 & 18.7 \\
            \midrule
            \rowcolor{gray!10}
            \textbf{Total}       & \textbf{4,938} & \\
            \bottomrule
            \end{tabular}}
            \end{minipage}
        \end{table}

        \clearpage

        \Cref{tab:supp_answer_dist} examines the answer option distribution. While most categories exhibit a mild bias toward option A (26--31\%), the overall distribution remains reasonably balanced across A--D. A notable exception is Global Narrative, where 44.8\% of answers fall outside the standard A--D options, as the contributing benchmarks (\eg, MMR-V~\cite{mmrv}) employ up to 12 answer choices. 
        
        \begin{table}[h]
        \centering
        \caption{Multiple-choice answer distribution across categories.}
        \label{tab:supp_answer_dist}
        \scalebox{0.85}{%
        \begin{tabular}{l rrrr r}
        \toprule
        \textbf{Category} & \textbf{A (\%)} & \textbf{B (\%)} & \textbf{C (\%)} & \textbf{D (\%)} & \textbf{Others (\%)} \\
        \midrule
          Fine-Grained Perception        & 30.7 & 23.5 & 23.6 & 18.3 & 3.9 \\
          Spatial World Understanding       & 26.2 & 28.6 & 27.7 & 17.0 & 0.4 \\
          Temporal Dynamics \& Tracking      & 28.1 & 23.6 & 22.2 & 19.1 & 7.0 \\
          Causality \& Logical Reasoning       & 25.7 & 21.8 & 19.6 & 16.3 & 16.6 \\
          Global Narrative       & 16.9 & 13.6 & 11.9 & 12.7 & 44.8 \\
        \midrule
          \rowcolor{gray!10}
          \textbf{Overall} & 26.6 & 23.2 & 21.9 & 17.7 & 10.6 \\
        \bottomrule
        \end{tabular}}
        \end{table}

        \Cref{tab:supp_benchmark_category} shows the relationship between benchmark characteristics and challenge categories. 
        While some benchmarks are dominated by temporal reasoning (\eg, TVBench~\cite{tvbench}), others emphasize spatial reasoning (\eg, VSI-Bench~\cite{vsibench}) or narrative reasoning (\eg, Video-Holmes~\cite{videoholmes} and MMR-V~\cite{mmrv}).
        This cross-benchmark analysis suggests that Video-Oasis aggregates complementary challenges from diverse benchmarks into a unified evaluation framework.

        \begin{table}[h]
        \centering
        \footnotesize
        \setlength{\tabcolsep}{3pt}
        \renewcommand{\arraystretch}{1.0}
        
        \caption{Per-benchmark distribution of samples across video-native challenges.}
        \label{tab:supp_benchmark_category}
        \begin{tabular}{lrrrrrr}
        \toprule
        \textbf{Benchmark} &
        \makecell{\textbf{Fine}\\\textbf{Percep.}} &
        \makecell{\textbf{Spatial}\\\textbf{World}} &
        \makecell{\textbf{Temporal}\\\textbf{Dynamics}} &
        \makecell{\textbf{Causal}\\\textbf{Reason.}} &
        \makecell{\textbf{Global}\\\textbf{Narrat.}} &
        \textbf{Total} \\
        \midrule
        EgoSchema~\cite{egoschema}          & 0 & 1 & 40 & 25 & 52 & 118 \\
        ImplicitQA~\cite{implicitqa}        & 17 & 270 & 39 & 25 & 5 & 356 \\
        VSI-Bench~\cite{vsibench}           & 0 & 1,151 & 399 & 0 & 0 & 1,550 \\
        \midrule
        TVBench~\cite{tvbench}              & 0 & 111 & 1,063 & 0 & 0 & 1,174 \\
        VCR-Bench~\cite{vcrbench}           & 2 & 6 & 197 & 45 & 5 & 255 \\
        RTV-Bench~\cite{rtvbench}           & 741 & 143 & 679 & 357 & 0 & 1,920 \\
        \midrule
        Video-Holmes~\cite{videoholmes}     & 1 & 0 & 132 & 412 & 413 & 958 \\
        MINERVA~\cite{minerva}              & 65 & 30 & 715 & 83 & 8 & 901 \\
        MMR-V~\cite{mmrv}                   & 16 & 1 & 66 & 119 & 456 & 658 \\
        \midrule
        VideoMME~\cite{videomme}           & 62 & 12 & 399 & 60 & 100 & 633 \\
        MVBench~\cite{mvbench}              & 6 & 49 & 740 & 199 & 0 & 994 \\
        LVBench~\cite{lvbench}              & 97 & 10 & 513 & 72 & 48 & 740 \\
        LongVideoBench~\cite{longvideobench}& 78 & 2 & 458 & 5 & 2 & 545 \\
        MLVU~\cite{mlvu}                    & 22 & 3 & 191 & 12 & 3 & 231 \\
        \midrule
        \rowcolor{gray!10}
        \textbf{Total} & \textbf{1,107} & \textbf{1,789} & \textbf{5,631} & \textbf{1,414} & \textbf{1,092} & \textbf{11,033} \\
        \bottomrule
        \end{tabular}
        \end{table}
                
        \subsection{Extended Qualitative Examples: Video-Native Challenges} \label{sec:d.2}
        \label{subsec:supp_qualitative_examples}
\begin{figure*}[ht!]
          \centering
          \includegraphics[width=0.9\linewidth]{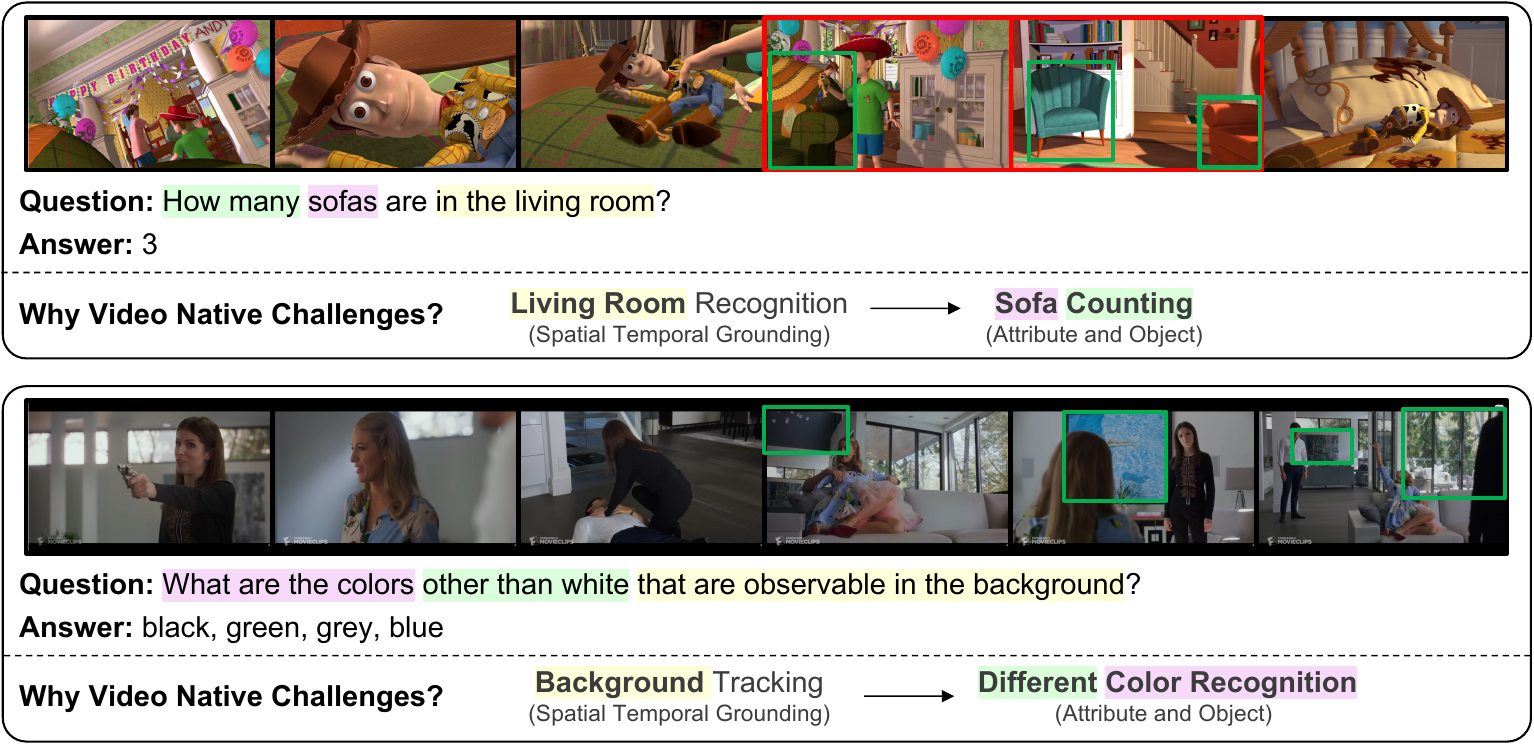}
          \caption{Qualitative examples of Fine-Grained Perception Challenges.} \vspace{-4mm}
          \label{fig:fig_sup_perception}
        \end{figure*}

        \noindent\textbf{Fine-Grained Perception} requires recognizing subtle visual cues that must be integrated across space and time.
        In the upper example of \Cref{fig:fig_sup_perception}, sofas are only partially visible from different viewpoints, requiring the model to combine fragmented evidence to infer the correct count.
        In the lower example, the model must distinguish non-white background colors in visually cluttered scenes. \vspace{-4mm}

\begin{figure*}[ht!]
          \centering
          \includegraphics[width=0.9\linewidth]{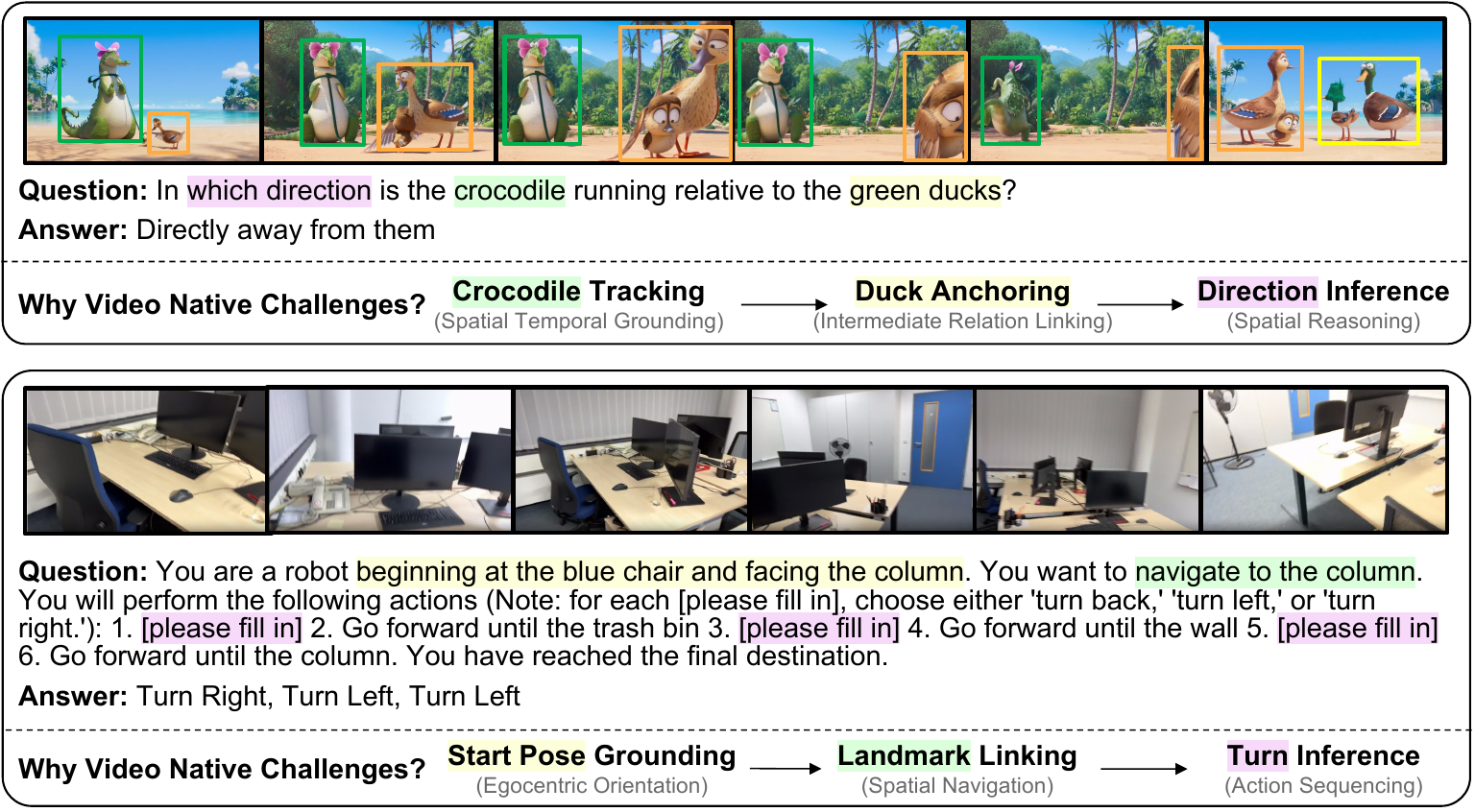}
          \caption{Qualitative examples of Spatial World Understanding Challenges.} \vspace{-4mm}
          \label{fig:fig_sup_spatial}
        \end{figure*}

        \noindent\textbf{Spatial World Understanding} requires integrating multi-view evidence across frames to infer spatial relations such as relative position, geometry, and motion. 
        In the upper example of \Cref{fig:fig_sup_spatial}, the task requires the model to infer the crocodile’s direction relative to the green ducks by linking spatial relations through the brown ducks. 
        In the lower example, reasoning necessitates navigation using the robot’s orientation and nearby landmarks.\vspace{-4mm}
    
\begin{figure*}[ht!]
          \centering
          \includegraphics[width=0.9\linewidth]{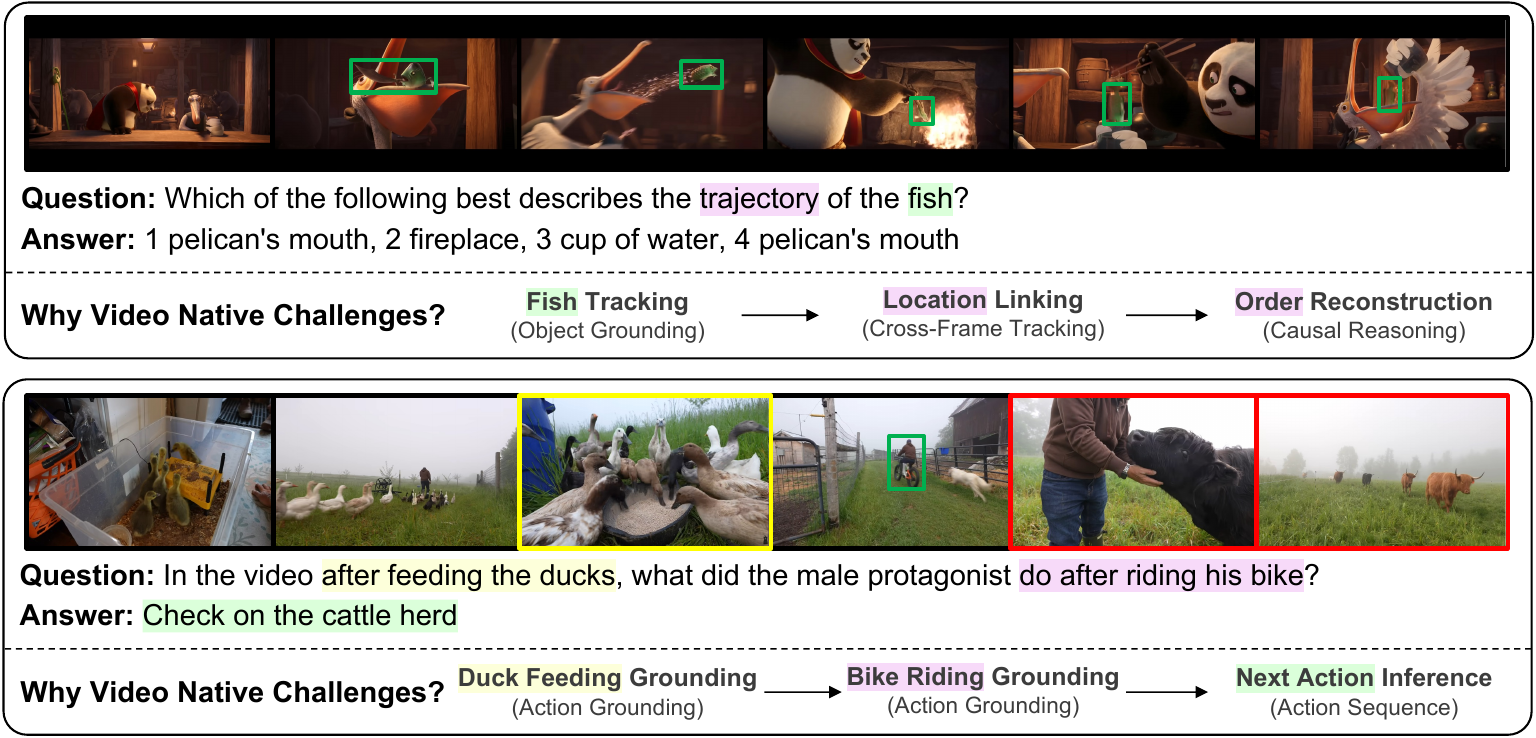}
          \caption{Qualitative examples of Temporal Dynamics \& Tracking Challenges.} \vspace{-4mm}
          \label{fig:fig_sup_temporal}
        \end{figure*}

        \noindent\textbf{Temporal Dynamics \& Tracking} requires reasoning over temporally ordered evidence, where the answer depends on chronological sequence rather than isolated frame matching. 
        In the upper example of \Cref{fig:fig_sup_temporal}, the model must reconstruct the trajectory of the fish across distinct positions in the correct temporal order. 
        In the lower example, solving the question requires establishing the correct temporal ordering of the protagonist's actions to identify the action following the bike ride. \vspace{-4mm}
        
\begin{figure*}[ht!]
          \centering
          \includegraphics[width=0.9\linewidth]{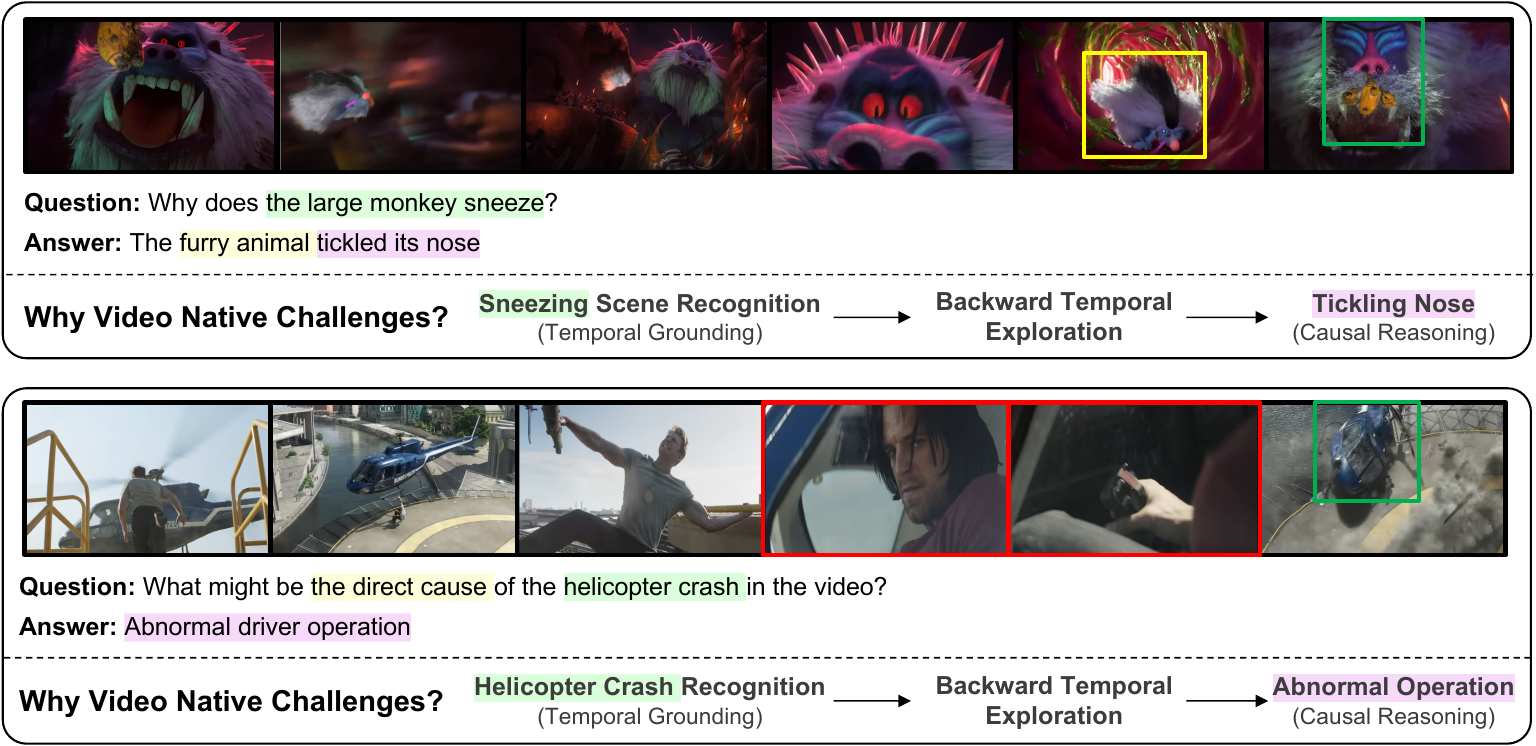}
          \caption{Qualitative examples of Causality \& Logical Reasoning Challenges.} 
          \label{fig:fig_sup_causal}
        \end{figure*}

        \noindent\textbf{Causality \& Logical Reasoning} requires inferring why events occur, rather than merely describing what happens, by reasoning about hidden causes and unobserved intentions. 
        In the upper example of \Cref{fig:fig_sup_causal}, the model needs to identify the tickling action as the cause of the sneeze, rather than simply noting their temporal sequence. 
        In the lower example, solving the question requires attributing the helicopter crash to abnormal driver operation by reasoning over the preceding events. \vspace{-4mm}
        
\begin{figure*}[ht!]
          \centering
          \includegraphics[width=0.9\linewidth]{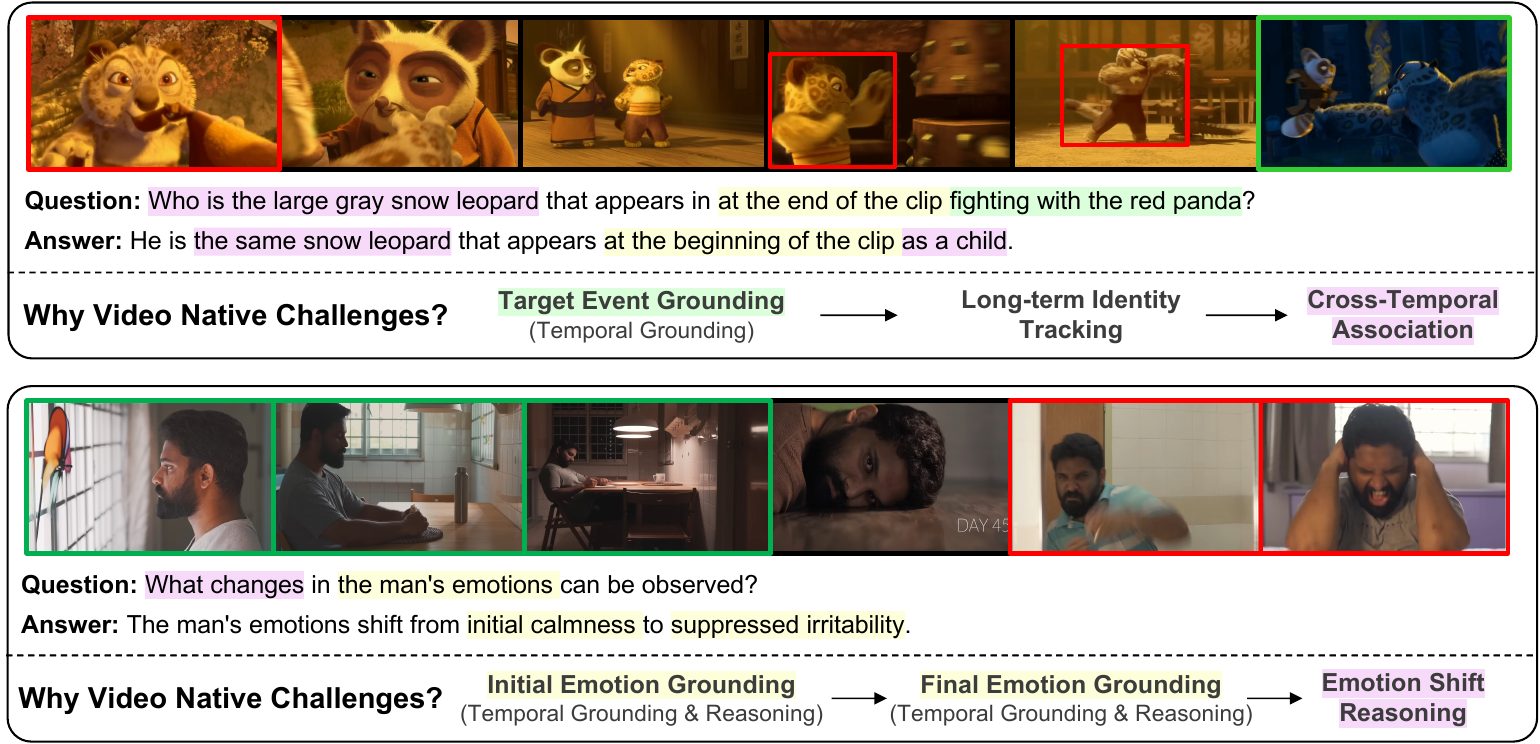}
          \caption{Qualitative examples of Global Narrative Challenges.} \vspace{-6mm}
          \label{fig:fig_sup_global}
        \end{figure*}

        \noindent\textbf{Global Narrative} requires aggregating dispersed events across the full timeline to capture narrative developments that only become clear over the course of the video. In the upper example of \Cref{fig:fig_sup_global}, the model must connect the snow leopard shown as a child at the beginning to the adult character at the end. In the lower example, it must infer the man's emotional shift from calmness to irritability by integrating behavioral and contextual cues across scenes. \vspace{-4mm}

    \subsection{Video-Native Challenge Annotation} \label{sec:d.3}
        For taxonomy construction, we use Gemini-2.5-Pro~\cite{gemini25} to derive candidate challenge clusters from source-benchmark metadata; the prompt template is shown in \Cref{tab:prompt_taxonomy}.
        For category assignment, we use an ensemble of five proprietary LLMs: OpenAI o3, OpenAI o4-mini, GPT-4o, GPT-5-mini, and GPT-5~\cite{o4mini,gpt4o,gpt5}.
        Each model receives the question, answer options, and category definitions, and predicts a single category label; the prompt template is shown in \Cref{tab:prompt_categorization}.
        We further examine whether the annotation results are robust to the choice of labeling models.
        While the main ensemble is based on OpenAI models, we additionally evaluate alternative ensembles that combine other closed-source models with open-source Video-LLMs.
        As shown in \Cref{tab:relabel_overlap}, these alternative ensembles produce highly consistent labels with the main ensemble, with overlaps above 93\%.
        This supports the robustness of the majority-voting procedure for category annotation.
        
        \begin{table}[h]
            \centering
            \scriptsize
            \caption{Robustness of category labeling under alternative model ensembles.}
            \label{tab:relabel_overlap}
            \resizebox{\columnwidth}{!}{%
            \begin{tabular}{l l l l l | c}
            \toprule
            \textbf{Model 1} & \textbf{Model 2} & \textbf{Model 3} & \textbf{Model 4} & \textbf{Model 5} & \textbf{Overlap} \\
            \midrule
            GPT-5-mini & OpenAI o3 & OpenAI o4-mini & Gemini-2.5-flash & Gemini-3-Flash & 95.8\% \\
            GPT-5-mini & OpenAI o3 & Gemini-3-flash & Claude Haiku 4.5 & Kimi-K2.0 & 93.7\% \\
            InternVL-3.5-8B & Eagle2.5-8B & OpenAI o3 & Claude Haiku 4.5 & Kimi-K2.0 & 93.2\% \\
            \bottomrule
            \end{tabular}%
            }
        \end{table}
    
        \begin{table}[t!] \centering
        \begin{tcolorbox}[colback=gray!5, colframe=gray!50, arc=2mm, boxrule=0.5pt, left=1.5mm, right=1.5mm, top=1mm, bottom=1mm] \footnotesize
        \textbf{System Prompt:} \\
        \vspace{1mm}
        We are currently conducting a meta-analysis of evaluation metrics from various video understanding benchmarks to construct a new integrated category taxonomy that can measure the pure `spatiotemporal understanding' of VLMs (Video Large Language Models). Below, we provide a list of dozens of sub-tasks used by these 14 existing benchmarks. Based on this data, please derive a classification criteria according to the following conditions. \\
        
        \textbf{[Input Data]} \\
        \textit{(A list of evaluation categories and sub-tasks associated with the surviving QA pairs from 14 benchmarks, which have already passed our rigorous shortcut-filtering process:)}
        \begin{itemize}[leftmargin=*, nosep]
            \item \textbf{ImplicitQA}: \texttt{Relative Depth and Proximity}, \texttt{Inferred Counting}, \texttt{Causal and Motivational Reasoning}, \dots
            \item \textbf{MVBench}: \texttt{Action Sequence}, \texttt{object interaction}, \texttt{counterfactual inference}, \dots
            \item \textit{\dots (Remaining task categories from 11 other benchmarks omitted for brevity)}
        \end{itemize}
        \vspace{1mm}
        
        \textbf{[Constraints \& Guidelines]}
        \begin{enumerate}[leftmargin=*, nosep]
            \item \textbf{Shortcut Filtering}: Analyze the provided list of existing sub-tasks and first isolate the tasks that can be solved via unintended shortcuts---such as relying on a single static frame (e.g., static object recognition, simple OCR) or pure linguistic priors. Set these aside to clearly delineate the boundary of true video understanding.
            \item \textbf{Taxonomy for Video Understanding}: Focusing solely on the remaining tasks that strictly mandate spatiotemporal dependencies and dynamic reasoning, group them based on the cognitive abilities they require. By abstracting these surviving tasks, establish a new, definitive category taxonomy that a ``genuine video understanding benchmark'' must possess.
        \end{enumerate}
        \vspace{1mm}
        
        \textbf{[Output Constraints]}
        \begin{itemize}[leftmargin=*, nosep]
            \item \textbf{STRICT FORMATTING}: Do not output any preambles, introductory text, postambles, or explanations of your filtering process (e.g., do not show ``Phase 1: Shortcut Filtering Analysis'').
            \item \textbf{NO EXTRA DETAILS}: Do not list the mapped sub-tasks, core abilities, or add any extra bullet points.
            \item \textbf{LENGTH LIMIT}: The definition must be highly concise, limited to a maximum of 3 sentences.
        \end{itemize}
        \vspace{1mm}
        
        \textbf{[Output Format]} \\
        Provide only the resulting N categories. You must strictly follow the exact structure below and output nothing else: \\
        
        \texttt{< Category Name >} \\
        \texttt{< Definition > : A clear academic definition of the category. Maximum 3 sentences.}
        \end{tcolorbox}
        \caption{Prompt template used to derive candidate video-native challenge clusters from source-benchmark metadata.}
        \label{tab:prompt_taxonomy}
        \vspace{-5mm}
        \end{table}

        \begin{table}[t!] \centering
        \begin{tcolorbox}[colback=gray!5, colframe=gray!50, arc=2mm, boxrule=0.5pt, left=2mm, right=2mm, top=2mm, bottom=2mm] \footnotesize
        \textbf{System Prompt:} \\
        You are an expert data annotator for a highly rigorous Long Video Understanding (LVU) dataset. \\
        
        \textbf{[Core Philosophy of Video-Oasis]} \\
        Existing long video benchmarks often include flawed questions solvable via a single frame or text alone (question \& options only / summaries), failing to evaluate genuine video understanding. To address this, our work rigorously filters out questions that can be answered using several frames, audio-only (STT), or text-only reasoning. Consequently, we retain only strictly video-dependent questions that mandate direct observation and synthesis of spatiotemporal context and temporal flow, thereby restoring the true essence of video capability evaluation. \\
        
        \textbf{[Task]} \\
        Guided by the philosophy above, your task is to classify a given [Question] and its [Options] into one of the predefined [Categories]. \\
        You must evaluate the fundamental cognitive mechanism and the level of visual and temporal context required to answer the question, always ensuring that the classification reflects the true video-dependent nature of the problem. You must deduce this solely from the text, as no actual video is provided. \\
        
        \textbf{[Question]} \\
        \{question\} \\
        
        \textbf{[Options]} \\
        \{options\} \\
        
        \textbf{[Categories]} \\
        \{categories\} \\
        
        \textbf{[Instruction]} \\
        Analyze the question and options, and assign it to the single most appropriate category. \\
        Output ONLY the single uppercase letter corresponding to your chosen category. \\
        Do not include any explanations, punctuation, or additional text. Your output must be exactly one character: `A', `B', `C', `D', `E', 
        \dots
        \end{tcolorbox}
        \caption{Prompt template used to assign each surviving QA pair to one of the defined video-native challenge categories.}
        \label{tab:prompt_categorization}
        \vspace{-8mm}
        \end{table}

        \clearpage

\section{Reproduction Results} \label{sec:e}

    Various methods are evaluated using Video-Oasis or under the video-native challenge setting. 
    To ensure experimental reliability, we provide a validation by comparing official benchmark scores with our reproduced results.
    For Video-LLMs, we fix the frame sampling protocol to a maximum of 128 frames at 1 fps for consistency across models.
    Although this differs from the official benchmark settings~\cite{eagle25,qwem3_vl,longvilar1}, which typically use more frames (\eg 512 or 2048 frames), the reproduced results remain within a reasonable margin of the reported scores.
    For agentic methods, we replace the reasoning models used in VideoTree~\cite{videotree} and STAR~\cite{videotool} with a more recent model, GPT-5-mini~\cite{gpt5}.
    Under this setting, the reproduced results achieve performance comparable to or higher than the reported results, as shown in \Cref{tab:reproduce}.

    \begin{table}[h]
    \centering
    \caption{Reproduction results on LongVideoBench and VideoMME. For each comparison, higher scores are highlighted in \textbf{bold}, while lower scores are \underline{underlined}.}
    \label{tab:reproduce}
    \small
    \begin{tabular}{lcccc}
    \toprule
    \multirow{2}{*}{\textbf{Model}} & \multicolumn{2}{c}{\textbf{LongVideoBench}~\cite{longvideobench}} & \multicolumn{2}{c}{\textbf{VideoMME}~\cite{videomme}} \\
    \cmidrule(lr){2-3} \cmidrule(lr){4-5}
     & Official & Reproduced & Official & Reproduced \\
    \midrule
    \rowcolor[gray]{0.9} \multicolumn{5}{l}{\textit{Open-Source Video-LLMs}} \\
    Qwen2.5-VL (7B)~\cite{qwen25vl} & - & 59.7 & \underline{65.1} & \textbf{65.3} \\
    Eagle2.5 (8B)~\cite{eagle25} & - & 68.0 & \textbf{72.5} & \underline{71.0} \\
    $\text{Qwen3-VL}_\text{Instruct}$ (8B)~\cite{qwem3_vl} & - & 64.7 & \textbf{71.4} & \underline{69.7} \\
    $\text{Qwen3-VL}_\text{Thinking}$ (8B)~\cite{qwem3_vl} & - & 63.2 & \textbf{71.8} & \underline{68.4} \\
    $\text{VideoAuto-R1}_\text{Qwen2.5}$ (7B)~\cite{videoautor1} & - & 61.2 & \textbf{67.3} & \underline{67.0} \\
    $\text{VideoAuto-R1}_\text{Qwen3}$ (8B)~\cite{videoautor1} & - & 65.4 & \textbf{71.7} & \underline{69.4} \\
    InternVL-3 (8B)~\cite{internvl3} & \underline{58.8} & \textbf{59.5} & \textbf{66.3} & \underline{65.4} \\
    InternVL-3.5 (8B)~\cite{internvl35} & \textbf{62.1} & \underline{59.2} & \textbf{66.0} & \underline{63.9} \\
    Video-R1 (7B)~\cite{videor1} & - & 60.0 & \underline{61.4} & \textbf{62.4} \\
    LongViLA-R1 (7B)~\cite{longvilar1} & \underline{58.0} & \textbf{59.0} & \textbf{65.1} & \underline{64.0} \\ \hline
    \rowcolor[gray]{0.9} \multicolumn{5}{l}{\textit{Agentic Methods}} \\
    VideoTree ($\text{GPT-5}_\text{{mini}}$)~\cite{videotree} & - & 56.1 & \underline{54.2} & \textbf{62.7} \\
    STAR ($\text{GPT-5}_\text{{mini}}$)~\cite{videotool} & \underline{57.2} & \textbf{63.0} & \underline{70.0} & \textbf{71.3} \\
    \bottomrule
    \end{tabular}
    \end{table}



\end{document}